\documentclass{ieeeaccess}

\usepackage{comment}
\usepackage{placeins}
\usepackage{xcolor}

\usepackage{cite}
\usepackage{amsmath,amssymb,amsfonts}
\usepackage{algorithmic}
\usepackage{algorithm}
\usepackage{graphicx}
\usepackage{textcomp}
\usepackage{subfigure}
\usepackage{multirow}
\usepackage{booktabs}
\usepackage{float}
\usepackage{cite}
\usepackage{amsmath,amssymb,amsfonts}
\usepackage{graphicx}
\usepackage{textcomp}
\usepackage{subfigure}
\usepackage{multirow}
\usepackage{booktabs}
\usepackage{float}
\usepackage{amssymb}
\newcommand{\xmark}{\ding{55}}
\usepackage{pifont}
\def\BibTeX{{\rm B\kern-.05em{\sc i\kern-.025em b}\kern-.08em
    T\kern-.1667em\lower.7ex\hbox{E}\kern-.125emX}}
\begin{document}
\history{Date of publication xxxx 00, 0000, date of current version xxxx 00, 0000.}
\doi{10.1109/ACCESS.2023.0322000}

\title{Long-term Human Participation Assessment In Collaborative Learning Environments Using Dynamic Scene Analysis}
\author{\uppercase{Wenjing Shi}\authorrefmark{1}, \IEEEmembership{Member, IEEE},
\uppercase{Phuong Tran}\authorrefmark{1},
\uppercase{Sylvia Celed\'on-Pattichis}\authorrefmark{2},
and \uppercase{Marios S. Pattichis}\authorrefmark{1}, \IEEEmembership{Senior Member, IEEE}}

\address[1]{Image and Video Processing and Communications Lab,
Dept. of Electrical and Computer Engineering,
University of New Mexico, Albuquerque, NM 87131 USA (e-mail: ntswj1@gmail.com, pattichi@unm.edu, pnt204@unm.edu)}
\address[2]{
Department of Curriculum and Instruction,
The University of Texas at Austin,
Austin, Texas 78712-1293
(e-mail: sylvia.celedon@austin.utexas.edu)}
\tfootnote{This work was supported in part by the National Science Foundation under Grant No. 1949230, No. 1842220, and No. 1613637.}

\markboth
{W. Shi \headeretal: Long-term Human Participation Assessment In Collaborative Learning Environments}
{W. Shi \headeretal: Long-term Human Participation Assessment In Collaborative Learning Environments}

\corresp{Corresponding author: Prof. Marios S. Pattichis (e-mail: pattichi@unm.edu).}

\begin{abstract}
The paper develops datasets and methods to assess
       student participation in real-life collaborative learning environments.       
In collaborative learning environments, students
       are organized into small groups where they
       are free to interact within their group.	
Thus, students can move around freely causing
       issues with strong pose variation,
       move out and re-enter the camera scene,
       or face away from the camera.
	
We formulate the problem of assessing student participation
       into two subproblems: 
       (i) student group detection against strong background interference
       from other groups, and (ii) dynamic participant tracking within the group.
A massive independent testing dataset of 
      12,518,250 student label instances, of total duration of 
      21 hours and 22 minutes of real-life videos,
      is used for evaluating the performance of our proposed
      method for student group detection.
The proposed method of using multiple
      image representations is shown to perform
      equally or better than YOLO on all video instances.
Over the entire dataset, the proposed method achieved
      an F1 score of 0.85 compared to 0.80 for YOLO.    
     
Following student group detection, the paper presents
      the development of a dynamic participant tracking system
      for assessing student group participation through long video sessions.
The proposed dynamic participant tracking system is shown
     to perform exceptionally well, missing a student in just 
     one out of 35 testing videos.
In comparison, a state-of-the-art method fails to track
     students in 14 out of the 35 testing videos.      
The proposed method achieves 82.3\% accuracy on an 
      independent set of long, real-life collaborative videos.
\end{abstract}


\titlepgskip=-21pt

\maketitle

\section{Introduction}
\label{sec:introduction}
Classroom video analysis requires the development
       of robust image processing methods that
       can work in very challenging environments.
In this paper, we study methods for student group detection,
        student recognition, and assessing student participation
        under challenging occlusions and student movement.
We demonstrate our methods on classroom videos
        that were collected by the Advancing Out-of-school Learning in 
        Mathematics and Engineering (AOLME) project.
        
\begin{figure}[t!]
    \centering
     \subfigure[Example with total occlusion.]
 {
        \includegraphics[width=0.45\textwidth]{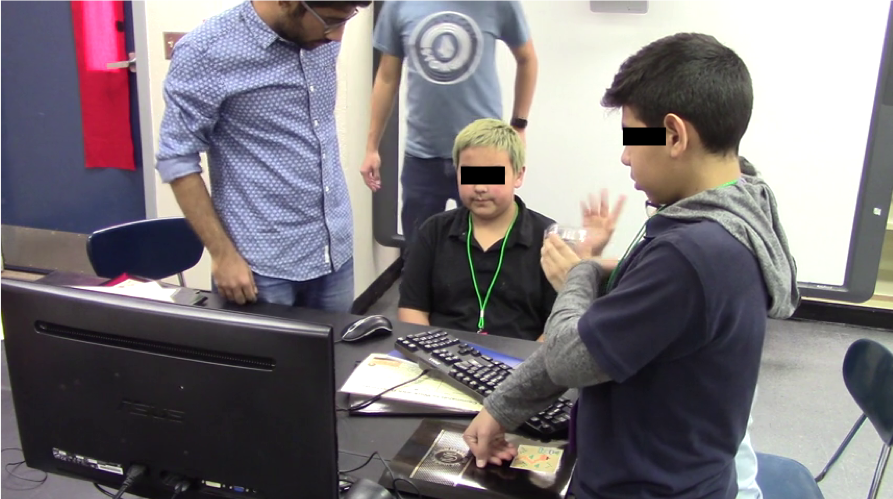}
        %\label{fig:second_sub}
    }\\
    \subfigure[Example with partial occlusion, complicated background, and multiple activities.]
    {
        \includegraphics[width=0.45\textwidth]{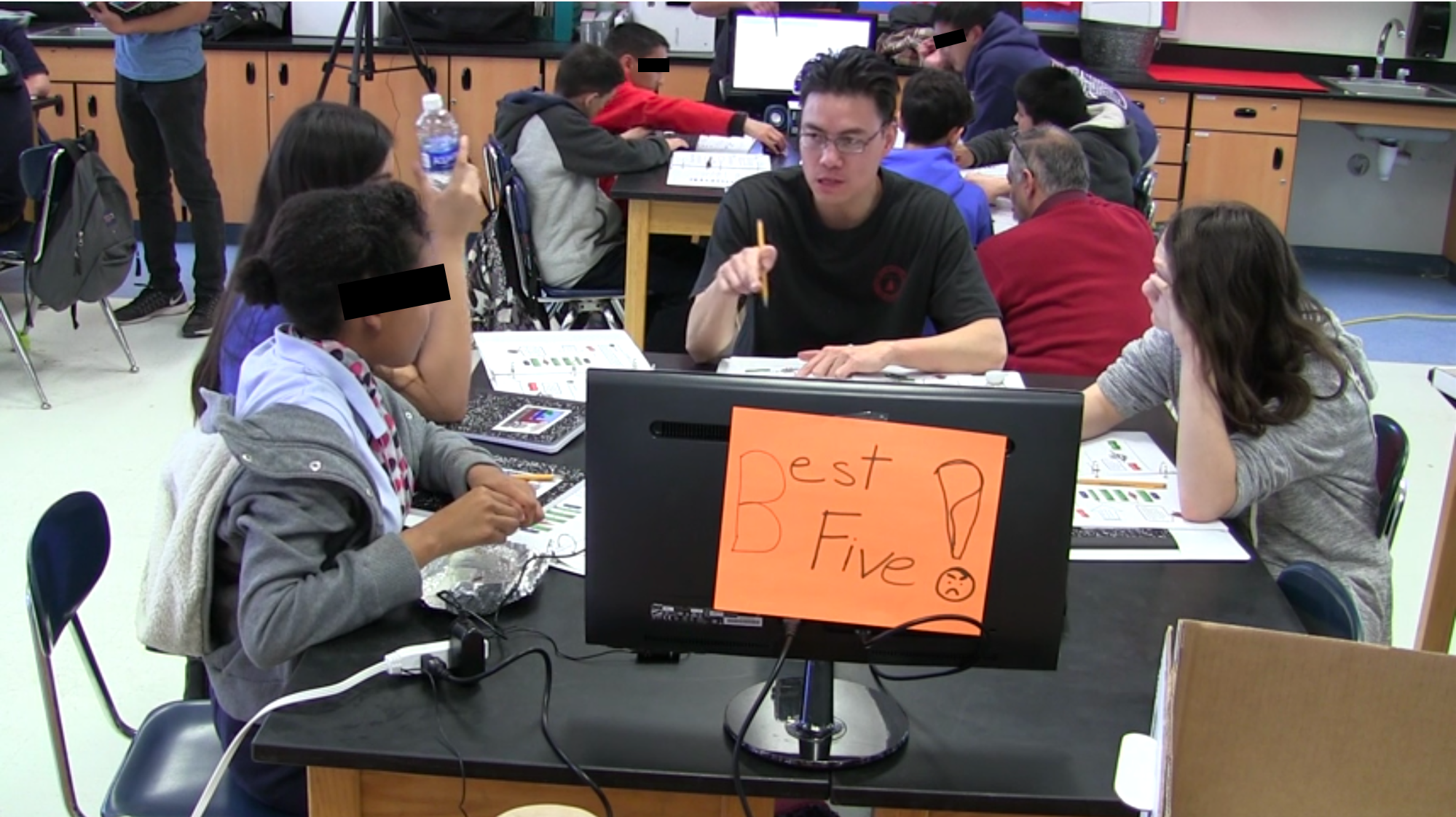}
        %\label{fig:second_sub}
    }

    \caption{Examples of the challenges associated with developing methods for assessing student participation based on the AOLME datasets.}
    \label{fig:aolme_examples}
\end{figure}

AOLME videos were recorded in actual student classrooms as demonstrated
         in Fig. \ref{fig:aolme_examples}. 
The classroom is organized into
        several groups of students, where multiple student groups can appear
        in a single video (see Fig. \ref{fig:aolme_examples}(b)).
We use a single video camera for each group.
Thus, our first task is to develop methods for
         student group detection, by detecting
         the students that are closest to the camera.
As it is clear from Fig. \ref{fig:aolme_examples},
         students need to be detected from multiple angles.
Furthermore, there are significant issues with both
         partial and full occlusions.         
Students can be active participants
         while they remain partially or fully occluded.
Thus, in order to properly assess student participation,
          we need to develop effective methods to
          deal with occlusions.         
In addition, we also need to deal with
        significant student movements in and out of the frame
        (see Fig. \ref{fig:aolme_examples}(a)).          
Furthermore, AOLME video sessions are very long,
        ranging from 45 to 90 minutes each.
As a result, we need to keep track of student
        participation throughout the long video sessions.
We provide an extensive comparison of the 
       AOLME video dataset against other datasets
       in section \ref{sec:datasets}.

%\subsubsection{AOLME datasets}

The unique challenges associated with processing the AOLME
       dataset require that we consider the development
       of new approaches.
Furthermore, due to the need to process large video datasets,
       we require the development of fast methods.
Specifically, we need to integrate person detection,
       face recognition, and tracking under occlusion.
These methods have to be integrated into a video analysis
      system that supports the long 
      durations of the AOLME video sessions.      
We use the term Dynamic Participant Tracking (DPT) to refer
      to our approach.
DPT processes the time history of group detections from 
      individual video frames to determine a state for each student
      participant (e.g., occluded, inside frame, outside frame, 
      inside and outside, and unknown).
Furthermore, DPT processes a sequence of frames to determine
      transitions from state to state.

We provide a summary of related methods that we have
      adopted for our DPT.
Due to its speed, we adopted the use of YOLO for person tracking.
For face recognition, we adopt the use of  
     the InsightFace system \cite{deng2019arcface} 
     that  has been tested on a large number of camera-facing image datasets
    and a variety of loss function models.
We will provide more details on related background methods in
    Section \ref{fig:edge_case}.

Here, we provide a brief summary of methods that have been
       recently developed to track objects under occlusion.
We note the use of  
         a correlation filter in \cite{chen2020augmented},
         a classifier approach in  \cite{dong2016occlusion}, 
         and convolutional neural networks in \cite{yuan2020scale}. 
More recently, a geometric approach  has been developed in 
        \cite{nasseri2021simple} and \cite{stadler2021improving}.
In  \cite{nasseri2021simple}, the authors proposed  
        a novel algorithm that addresses occlusion by using only the 
        location and size of detection bounding boxes. 
The algorithm, termed
         Simple Online and Real-time Tracking with Occlusion Handling (SORT\_OH \cite{nasseri2021simple})
        can predict occlusions and re-identify lost targets. 
This paper uses both MOT16/17 datasets for pedestrian tracking and achieved 
       state-of-the-art results for online tracking algorithms. 
We will provide comparisons of our proposed approach
        against  Simple Online and Real-time Tracking with Occlusion Handling (SORT\_OH \cite{nasseri2021simple})
       to demonstrate that we can achieve 
       significantly better performance on the AOLME dataset.

We claim four primary contributions.
First, we develop a system for student group detection using 
       multiple image representations.
As we document in our results, the use of multiple representations
       results in much better person detection.
Second, we develop a system for video face recognition for identifying the students
	          within the group.
Our video face recognition enables face recognition from different angles.
Third, we develop new methods for dynamic scene analysis system using DPT.
We demonstrate that the DPT provides much better results than SORT\_OH.
Fourth, we introduce the use of student participation maps for visualizing
       the results over long video sessions.

We note that we presented preliminary results 
          on group detection in conference publications:
          \cite{shi2016human},  \cite{shi2018robust},  \cite{shi2018dynamic},  
          \cite{shi2021person}, and
          video face recognition in \cite{tran2021facial}.
While we review these earlier methods for completeness,
          we note that the current paper describes
          training and testing on the complete system over
          much larger datasets.
Furthermore, the dynamic participant tracking methodology
        that is a primary focus of the current paper has never appeared
        in any previous publications.          
The paper also uses participant maps that were initially developed
        in \cite{jatla2023long} for tracking 
        student activities associated with hand movements
        (see \cite{jatla2021long}, \cite{teeparthi2021fast}).
Here, we note that the current paper does not involve any
       student activities that include hand movements.        
Overall, the complete system, including the dynamic scene analysis,
        has not been previously discussed in the literature.                       

We organize the rest of the paper into five additional sections.
In Section \ref{sec:datasets}, we provide
        a detailed description of the AOLME dataset
        and elaborate on its challenges as we compare
        against other datasets.
In Section \ref{sec:background},
         we provide detailed background 
         information.
We then describe our proposed methods
        in Section  \ref{sec:methods}.
The results are given in Section \ref{sec:results}.
We provide concluding remarks in
        Section \ref{sec:conclusion}.

\begin{table*}[h]
	\caption{\label{tbl: Tracking data comparison}
		 AOLME dataset uniqueness against common video datasets.
		 AOLME contains real-life recordings of actual classrooms
		 with significant challenges.}	
		\begin{center}
		\resizebox{\textwidth}{!}{
			\begin{tabular}{lcccccc} 
				% Title row: 
				%   Use center alignments for the table headings.
				\toprule % booktabs 
				
				\textbf{Features} & \textbf{MOT16/17\cite{MOT16}} & \textbf{OTB-2015\cite{WuLimYang13}} & \textbf{VOT2018\cite{VOT_TPAMI}} & \textbf{LaSOT\cite{fan2019lasot}} & \textbf{TAO\cite{dave2020tao}} & \textbf{AOLME} \\
				% \textbf{\begin{tabular}[c]{@{}l@{}}Multiple activities\\ in one video\end{tabular}} &
				% \textbf{\begin{tabular}[c]{@{}l@{}}Activities at\\ different scales\end{tabular}} \\
				
				\midrule 
				Various camera angles & \checkmark & \checkmark & \checkmark & \checkmark & \checkmark & \checkmark \\    \addlinespace%\midrule 
				
				Multiple objects \\and humans & \checkmark  &  \checkmark &  \checkmark & \checkmark & \checkmark & \checkmark \\    \addlinespace%\midrule 
				
				Diverse scales \\
				of activities & \checkmark & \checkmark & \checkmark & \checkmark & \checkmark & \checkmark \\   \addlinespace% \midrule 
				
				Complicated background & \checkmark & \checkmark & \checkmark & \xmark & \xmark & \checkmark \\    \addlinespace%\midrule 
				
				Complete occlusion & \checkmark & \checkmark & \checkmark & \xmark & \xmark & \checkmark \\    \addlinespace%\midrule 
				
				Multiple activities & \checkmark & \xmark & \checkmark & \xmark & \xmark & \checkmark \\   \addlinespace %\midrule 
				
				Humans are \\
				at the edge \\
				of the frame & \xmark & \xmark & \xmark & \checkmark & \checkmark & \checkmark \\     \addlinespace%\midrule 
				
				Specific group detection & \xmark & \xmark & \xmark & \xmark & \xmark & \checkmark \\    \addlinespace%\midrule 
				
				Long-term occlusion & \xmark & \xmark & \xmark & \xmark & \xmark & \checkmark \\    \addlinespace%\midrule 
				
				Tracking specific \\
				objects throughout \\
				the video & \xmark & \xmark & \xmark & \xmark & \xmark & \checkmark \\    \addlinespace%\midrule 
				
				Video length & < 1 min 25 secs       & <  2 min 9 secs & < 1 min 23 secs  & $\approx$ 83 secs 
				                         & $\approx$ 30 secs &  23 min 45 secs segments  \\             
				\bottomrule % booktabs  
			\end{tabular}}
		\end{center}
	\end{table*}

%%%%%%%%%%%%%%%%%%%%%%%%%%%%%%%%%
\section{AOLME Student Datasets}\label{sec:datasets}
We provide a comparison of the unique characteristics of the AOLME 
       dataset as compared against related video datasets in 
       Table \ref{tbl: Tracking data comparison}.
We begin with a summary of common datasets and then
        provide a summary of the characteristics that are unique
        to AOLME.

We begin with a summary of common datasets.
Large-scale Single Object Tracking (LaSOT \cite{fan2019lasot}) 
         is used for single-object tracking with an average video length of 
         approximately 83 seconds.
The Tracking Any Object (TAO \cite{dave2020tao}) has 2907 videos and 833 classes, 
        where each video only includes a single activity lasting around 30 seconds. 
Both LaSOT and TAO datasets  are characterized by simple backgrounds
        and partial, short-term occlusions.
In contrast, the AOLME video dataset is characterized by complex backgrounds
         with both partial and full longer-term occlusions.

The Visual Tracker Benchmark 2015 (OTB-2015 \cite{WuLimYang13}) 
     contains 100 video clips with various activities and different objects.
Example objects include humans and SUVs.
The entire OTB-2015 dataset contains 58,613 frames, 
    and each video only has one type of activity. 
In contrast, the AOLME dataset is significantly larger with
    far more complex activities.

There are 60 sequences in the Visual Object Tracking 2018 (VOT2018 
     \cite{VOT_TPAMI}) datasets at a frame rate of about 30 fps.
The total duration for the dataset is only 745.2 sec. 
Furthermore, unlike AOLME, as for OTB-2015 and VOT2018,
      the dataset does not contain multiple, overlapping activities.

For human tracking, the most commonly used datasets include
      Multiple Object Tracking 16/17 (MOT16/17 \cite{MOT16}).
The datasets cover short-term  and full occlusions.
However, unlike AOLME, each video lasts less than 90 seconds.

In summary,  common datasets share videos captured from 
        multiple video angles that can 
        include multiple objects and humans at diverse scales.
In contrast, AOLME is characterized by the 
       need to develop methods for specific group detection, long-term occlusions, 
       and the need to track specific objects over very long video segments.       
AOLME video sessions range from 45 to 90 minutes broken into
      shorter segments of 23 minutes and 45 seconds.       
Overall, the AOLME dataset  
        contains over 950 hours of video,
        collected over three different cohorts, 
        with each cohort including 1$\sim$3 curriculum levels.
Within each cohort, we collected 
        10$\sim$12 video sessions 
        of 10-20 students collaborating
        in small groups of 3 to 6 members.
Thus,  it is clear that we need to develop
         methods that  detect specific 
         groups of students and track them throughout
         each video.

We tackle the problem of assessing long-term
         student participation into three subproblems that
         include (i) student group detection, 
         (ii) student face recognition within the detected group, and
         (iii)  dynamic participant tracking.   
We develop separate training and testing datasets for
          each problem as summarized in 
          Table \ref{table:aolme_student_datasets}.
We use different video sessions for training and testing.          
At the end, we use final testing datasets
          for measuring the performance of the integrated system.           
In what follows, we provide detailed descriptions of 
          the different datasets used to develop our system.
          
\begin{table*}[!h]
\centering
\caption{\label{table:aolme_student_datasets}AOLME student datasets.}
\resizebox{\textwidth}{!}{
\begin{tabular}{|ll|l|ll|ll|l|}
\toprule
\multicolumn{2}{|c|}{\multirow{2}{*}{\textbf{Problem}}}                                                                                                                            & \multicolumn{1}{c|}{\multirow{2}{*}{\textbf{Method}}}                         
& \multicolumn{2}{c|}{\textbf{Training/Validation Dataset}}                                                                                                     & \multicolumn{2}{c|}{\textbf{Testing Dataset}}                                                                                                                   & \multicolumn{1}{c|}{\multirow{2}{*}{\begin{tabular}[c]{@{}c@{}}\textbf{Final Testing}\\ \textbf{Dataset}\end{tabular}}} \\ \cline{4-7}
\multicolumn{2}{|c|}{}                                                                                                                                                             & \multicolumn{1}{c|}{}                                                         & \multicolumn{1}{c|}{\begin{tabular}[c]{@{}c@{}}\textbf{Dataset}\\ \textbf{Source}\end{tabular}} & \multicolumn{1}{c|}{\textbf{Dataset}}                                & \multicolumn{1}{c|}{\begin{tabular}[c]{@{}c@{}}\textbf{Dataset}\\ \textbf{Source}\end{tabular}} & \multicolumn{1}{c|}{\textbf{Dataset}}                                  & \multicolumn{1}{c|}{}                                                                                          \\ \toprule
\multicolumn{1}{|l|}{\multirow{3}{*}{\begin{tabular}[c]{@{}l@{}}\textbf{Group} \\ \textbf{detection}\end{tabular}}} & \multirow{2}{*}{Face detection}                                       & YOLO                                                                          & \multicolumn{1}{l|}{\multirow{3}{*}{AOLME-G}}                                          & \begin{tabular}[c]{@{}l@{}}AOLME-GY1\\ (2,200 images)\end{tabular}   & \multicolumn{1}{l|}{\multirow{3}{*}{AOLME-G}}                                          & -                                                                      
& \multirow{3}{*}{\parbox{2cm}{AOLME-GT\\ 13 videos\\ 12,518,250 labels\\ 21 hours 22 min}}                                                                                      \\ \cline{3-3} \cline{5-5} \cline{7-7}
\multicolumn{1}{|l|}{}                                                                                     &                                                                       & \begin{tabular}[c]{@{}l@{}}Group face \\ classifier\end{tabular}              & \multicolumn{1}{l|}{}                                                                  & \begin{tabular}[c]{@{}l@{}}AOLME-GF1\\ 112,129 images\end{tabular} & \multicolumn{1}{l|}{}                                                                  & \begin{tabular}[c]{@{}l@{}}AOLME-GF2\\  28,032 images\end{tabular}    &                                                                                                                \\ \cline{2-3} \cline{5-5} \cline{7-7}
\multicolumn{1}{|l|}{}                                                                                     & \begin{tabular}[c]{@{}l@{}}Back-of-the-head \\ detection\end{tabular} & \begin{tabular}[c]{@{}l@{}}Group\\ back-of-the-head\\ classifier\end{tabular} & \multicolumn{1}{l|}{}                                                                  & \begin{tabular}[c]{@{}l@{}}AOLME-GB1\\ 45,568 images \end{tabular}  & \multicolumn{1}{l|}{}                                                                  & \begin{tabular}[c]{@{}l@{}}AOLME-GB2\\ 11,392 images \end{tabular}    &                                                                                                                \\ \midrule
\multicolumn{2}{|l|}{\textbf{Face  recognition}}                                                                                                                                   & Extended InsightFace                                                                   
& \multicolumn{1}{l|}{AOLME-FR}                                                           
& \begin{tabular}[c]{@{}l@{}}AOLME-FR1\\ 3,968 images\end{tabular}    & \multicolumn{1}{l|}{-}                                                                 & -                                                                      
& \begin{tabular}[c]{@{}l@{}}AOLME-DLT\\ 2h 21m 24s videos\end{tabular}                                         \\ \midrule
\multicolumn{2}{|l|}{\textbf{Dynamic Participant Tracking}}                                                                                                                        & \begin{tabular}[c]{@{}l@{}}Dynamic scene \\ analysis\end{tabular}             & \multicolumn{1}{l|}{-}                                                                 & -                                                                    & \multicolumn{1}{l|}{AOLME-D}                                                          & \begin{tabular}[c]{@{}l@{}}AOLME-DST\\ 17m 17s videos \end{tabular} 
& \begin{tabular}[c]{@{}l@{}}AOLME-DLT\\ 2h 21m 24s videos\end{tabular} \\ 
\bottomrule
\end{tabular}}
\end{table*}

% Please add the following required packages to your document preamble:
% \usepackage{multirow}
% Please add the following required packages to your document preamble:
% \usepackage{multirow}

\subsection{AOLME-G video dataset for student group detection}\label{sec:AOLME-G} 
The AOLME-G video dataset has 54 videos from 52 groups, covering two cohorts.
These videos will be used for group detection.
We use AOLME-G to generate separate datasets for 
       (i) training and validation: AOLME-GY1, AOLME-GF1, AOLME-GB1, and
       (ii) component testing datasets:  AOLME-GF2 and AOLME-GB2.
We then want to test the group detection system
      using the massive AOLME-GT dataset.
We provide separate descriptions for each dataset.      

\subsubsection{AOLME-GY1 for face detection training and validation}
We use 1000 faces and 1200 non-face images from student groups extracted from the AOLME-G dataset to train the YOLO face detector. 
Among the selected face images, we use 70\% of the images for training and 30\% for validation.
For each group, we identify the faces of each group member.

\subsubsection{AOLME-GF1 for group face detection training and validation}
The dataset is generated from the AOLME-G videos to train the group face classifier. 
The augmented dataset contained 
        56,045 group faces and 56,084 non-group face images. 
        We use 70\% of the dataset for training and 30\% for validation.

\subsubsection{AOLME-GB1 for training for back of the head detection}
For the back-of-the-head classifier, the dataset uses over 45,000 frames from AOLME-G videos. 
It contains 22,768 back-of-the-head images and 22,800 other images.

\subsubsection{AOLME-GF2  for group face detection testing}
The dataset is generated from AOLME-G videos for testing the group face classifier. 
The dataset contains  14,011 group faces and 14,021 non-group face images.
The numbers include seven-fold data augmentation performed using random rescaling, 
       cropping, rotating, and flipping.

\subsubsection{AOLME-GB2 for back of the head testing}
To test the back-of-the-head classifier, we used 
     5,710 back-of-the-heads and 5,682 others from the AOLME-G video dataset.

\subsubsection{AOLME-GT: A large dataset for final testing of group detection}\label{sec:AOLME-GT}
We test the group detection methodology with a set of 13 videos containing 12,518,250 student labels.
The student labels identify whether a student belongs to a group or not.
Overall, the combined duration of all of the
     AOLME-GT videos is 21 hours and 22 minutes.

\subsection{AOLME-FR video face recognition training}
The AOLME-FR video dataset is used for training the video face recognition algorithms.
These video images were sampled from 13 sessions that cover level 1 of cohorts 2 \& 3.
Overall, the combinations of training and testing videos are 4 hours long.

\subsubsection{AOLME-FR1 for training video face recognition}
Within AOLME-FR, we separate out the AOLME-FR1 dataset that consists of 
        3,968 images for identifying up to 42 students and student facilitators.
The dataset is used to generate face prototypes associated with each participant as described
        in the methodology.
Each prototype is resized to  112$\times$112 pixels.

\subsubsection{AOLME-D for system testing of video face recognition from raw input videos}
The AOLME-D video dataset has 13 different sessions of 1 to 1.5 hours each
       from urban and rural schools.
We use AOLME-D to derive a collection of short videos (AOLME-DST) and long videos (AOLME-DLT)
       for final system testing.         

\subsubsection{AOLME-DST: Short videos dataset for final system testing}\label{sec:AOLME-DST}
The AOLME-DST dataset is summarized in Table \ref{table:AOLME-DST}.
This diverse dataset contains a selection of short video samples that are used
        to test system performance under occlusion for 35 students from 5 groups.
The AOLME-DST dataset is designed to provide exhaustive testing in many
        different scenarios.
In the results, we will provide detailed results for each group.        

\subsubsection{AOLME-DLT: Long videos dataset for final system testing}\label{sec:AOLME-DLT}
The AOLME-DLT contains raw real-life videos as detailed in Table \ref{table:AOLME-DLT}.
The videos are broken into shorter videos that are 23 minutes and 45 seconds.
This final dataset will be used to test all aspects of our system using different groups and
        a diverse set of students.

\begin{table}[!h]
\caption{\label{table:AOLME-DST} AOLME-DST: Short-video dataset for entire system
	            testing of dynamic participant tracking.
	            The test dataset includes 35 videos and 35 different students.}
\begin{center}
\begin{tabular}{p{0.05\textwidth}p{0.05\textwidth}p{0.07\textwidth}p{0.07\textwidth}p{0.08\textwidth}} \toprule
\textbf{Group} & \textbf{Video}  & \textbf{Duration} & \textbf{Occlusion Frames} & \textbf{Occlusion Time}  \\ \toprule
\multirow{15}{*}{\parbox{2cm}{G1\\ 4 students}} 
                                   & {V1} & {20s}& {434} & {14.5s}  \\
                                   & {V2} &{20s}& {104} & {3.5s}    \\   
                                   & {V3} & {25s}& {188} & {6.3s}  \\
                                   & {V4} &{45s}& {221} & {7.4s}    \\
                                   & {V5} & {1m45s}& {805} & {26.8s}  \\
                                   & {V6} &{10s}& {23} & {0.7s}    \\
                                   & {V7} & {50s}& {189} & {6.3s}  \\
                                   & {V8} &{20s}& {5} & {0.2s}    \\
                                   & {V9} & {30s}& {164} & {5.5s}  \\
                                   & {V10} &{15s}& {222} & {7.4s}    \\
                                   & {V11} & {10s}& {56} & {1.9s}  \\
                                   & {V12} &{20s}& {339} & {11.3s}    \\
                                   & {V13} & {20s}& {315} & {10.5s}  \\
                                   & {V14} &{5s}& {29} & {1s}    \\
                                   & {V15} & {10s}& {28} & {1s}  \\ \midrule
\multirow{7}{*}{\parbox{2cm}{G2\\ 6 students}}   
                       & {V1} &{10s}& {15} & {0.5s}    \\  
	              & {V2} &{26s}& {291} & {9.7s}    \\                                                                      
                                   & {V3} &{32s}& {724} & {24.1s}    \\ 
                                   & {V4} &{10s}& {22} & {0.7s}    \\  
	              & {V5} &{20s}& {61} & {2s}    \\                                                                      
                                   & {V6} &{15s}& {152} & {5.1s}    \\  
                                   & {V7} &{35s}& {300} & {10s}    \\ \midrule
\multirow{6}{*}{\parbox{2cm}{G3\\ 5 persons}} 
                       & {V1} &{50s}& {90} & {3s}    \\  
	              & {V2} &{10s}& {119} & {4s}    \\                                                                      
                                   & {V3} &{5s}& {36} & {1.2s}    \\ 
                                   & {V4} &{10s}& {97} & {3.2s}    \\  
	              & {V5} &{35s}& {219} & {7.3s}    \\                                                                      
                                   & {V6} &{1m1s}& {975} & {32.5s}    \\  \midrule
\multirow{1}{*}{\parbox{2cm}{G4\\ 5 students}}  
                      & {V1} &{1m15s}& {1605} & {53.5s}    \\  
                      & ~\\ \midrule
\multirow{6}{*}{\parbox{2cm}{G5\\ 6 students}}   
                      & {V1} &{20s}& {142} & {4.7s}    \\  
	              & {V2} &{15s}& {13} & {0.4s}    \\                                                                      
                                   & {V3} &{2m26s}& {906} & {30.2s}    \\ 
                                   & {V4} &{32s}& {1} & {0.03s}    \\  
	              & {V5} &{10s}& {120} & {4s}    \\                                                                      
                                   & {V6} &{15s}& {219} & {7.3s}    \\ 
                                                               			
			\bottomrule
		\end{tabular}
	\end{center}
\end{table}

\begin{table}[!h]
\caption{\label{table:AOLME-DLT}
   AOLME-DLT: Long-video dataset for entire system testing of 
                        dynamic participant tracking.
   The dataset contains videos from 22 students.}
\begin{center}
\begin{tabular}{ccc}
    \toprule % booktabs     
   \textbf{Video}  & \textbf{No. of students in group} & \textbf{Duration} \\
%{p{0.1\textwidth}p{0.1\textwidth}p{0.2\textwidth}} \toprule
\midrule
	              {V1} & {4} & \multirow{6}{*}{{23 minutes 45 seconds}} \\
                                   {V2} & {2} &   \\  
                                   {V3} & {4} & \\
                                   {V4} & {5}  &  \\
                                   {V5} & {4} & \\
                                   {V6} & {3}&   \\                                                                                        
			\bottomrule
		\end{tabular}
	\end{center}
\end{table}

\subsection{Datasets for Dynamic Participant Tracking}
The ultimate goal of dynamic participant tracking is to 
         quantify student participation.
Thus, we need to know whether a specific
          student is present within a group.
Students are marked as present even if
          they do not appear in the frame due
          to occlusion.
Thus, in order to develop ground truth
          for dynamic participant tracking,
          we review the entire video from beginning to end
          to eliminate false negatives due to occlusion.
Furthermore, in most cases, students are partially
          occluded and 
          are free to move around while remaining close to the table.
In all such cases, we assume that the students are present.
We only mark students as not-present if they are completely missing
         from several video frames over several seconds.        
         
\begin{figure*}[t!]
    \centering
    \subfigure[The student is present appearing at an angle with partial occlusion.]
    {
        \includegraphics[width=0.48\textwidth]{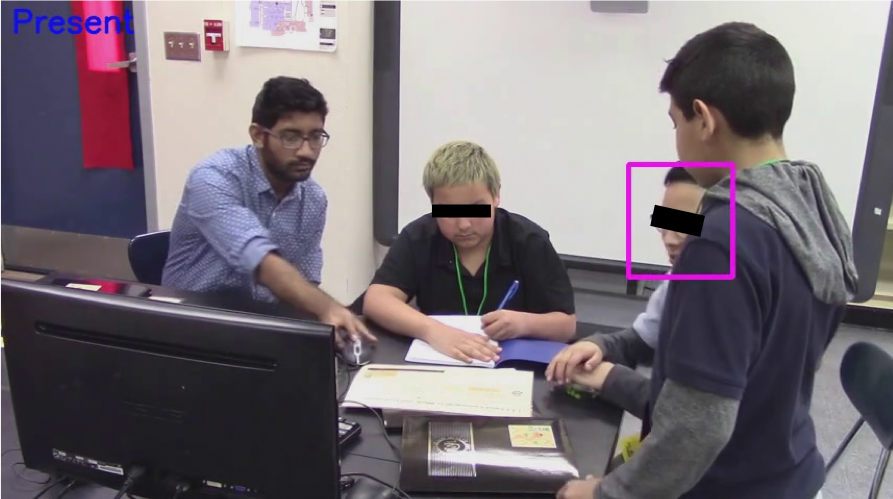}
        \label{fig:presented}
    }
    \subfigure[The student is present and fully occluded.]
    {
        \includegraphics[width=0.48\textwidth]{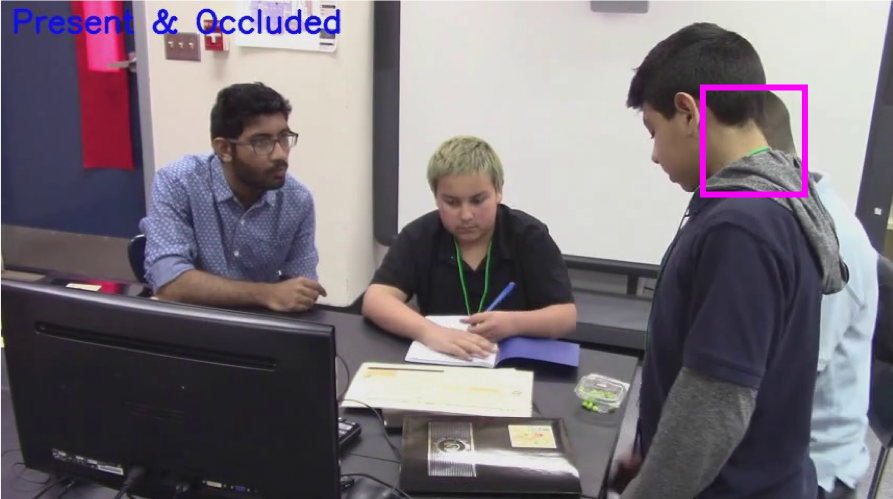}
        \label{fig:occluded}
    }
    \\
       \subfigure[The student is present at the edge of the frame.]
    {
        \includegraphics[width=0.48\textwidth]{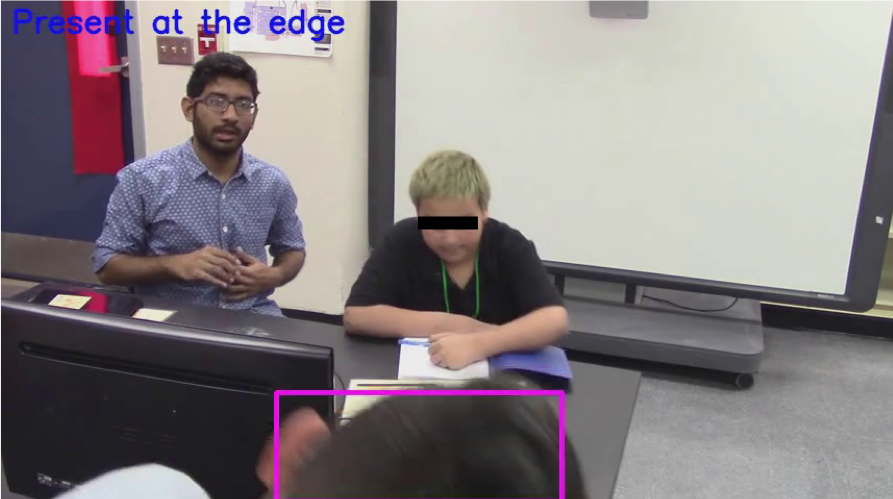}
        \label{fig:edge_2}
    }
     \subfigure[The student is present with his hand in the lower-right  edge of the frame.]
    {
        \includegraphics[width=0.48\textwidth]{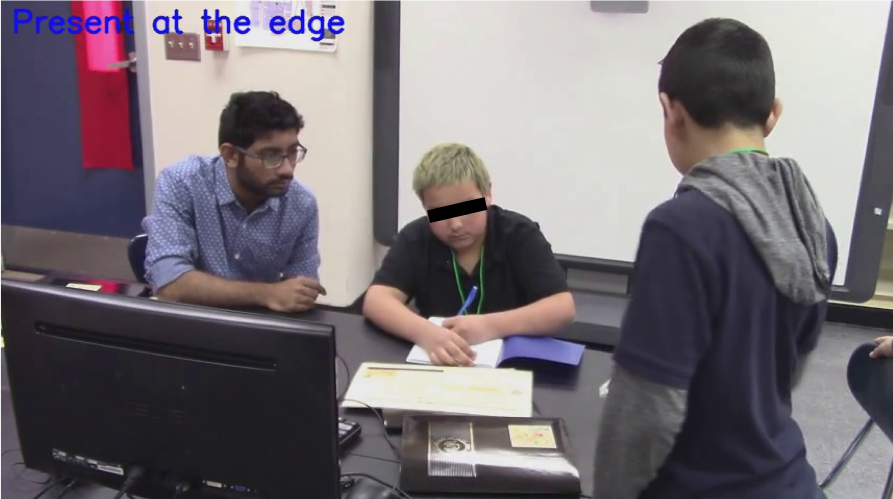}
        \label{fig:edge_1}
    }
    \caption{A simple example to demonstrate the issues for training and testing dynamic 
    participant tracking. In this example, we only show annotation for a single student per image.
    We note that there is no bounding box for the student in (d) because he is not visible. 
    For the training and testing datasets, in each frame, we mark all of the students for each group.}
    \label{fig:edge_case}
\end{figure*}

We present four occlusion examples in  Fig. \ref{fig:edge_case}.
In all cases, we mark the student as present.
Yet, the student is partially occluded in Fig. \ref{fig:presented},
         fully occluded in  Fig. \ref{fig:occluded},
        and at the edge of the frame in Fig. \ref{fig:edge_2}.
In Fig. \ref{fig:edge_1}, a small portion of his hand is visible 
        in the lower-right edge of the video frame.
        
We used the Matlab video labeler to mark the presence of each
       student in each frame of each video.
For each video frame, we carefully mark the locations of all students within each group.          

\subsubsection{AOLME-DST for system testing of dynamic participant tracking}\label{sub:AOLME-DST}
We perform both short-term and long-term testing of the ability of the system to perform
dynamic participant tracking.
For short-term testing, we use 35 short video segments ranging from 
10 seconds to 150 seconds long at a frame rate of 30 fps.
Overall, short-term testing consisted of 17 minutes and 17 seconds.
The video examples include occlusion of at-least one person
as detailed in Table \ref{table:AOLME-DST}.

\subsubsection{AOLME-DLT for system testing of dynamic participant tracking on long-duration videos}\label{sub:AOLME-DLT}
We use a second dataset to test our dynamic participant tracking system over long video segments.
Six long videos from different groups from the AOLME-CT video dataset are used to generate separate testing
videos for AOLME-DLT.
For long-term testing, each video is 23 minutes 45 seconds at a frame rate of 30 fps
with 3 to 5 recognizable persons per video as described in Table 
    \ref{table:AOLME-DLT}.
As for the short-term video dataset, we mark the location of every person in each video frame.

%%%%%%%%%%%%%%%%%%%%%%

\section{Background}\label{sec:background}

\subsection{Related work}
\subsubsection{Group Detection}
%\%\%\%\%\% Modify this part with different sentences.\%\%\%\%\%
We formulate the problem of group detection as a problem
     of detecting the students working together and sitting at the table
     nearest to the camera.
Beyond the classic problem of human detection, group detection
     requires that we detect humans at arbitrary angles, while
     facing the camera and also while looking away from the camera.
 In this subsection, we will summarize some related research done by our
     group, published in a conference paper, and outline the new research
     summarized in the current paper.

We reported on initial research of combining YOLO with AM-FM representations
      for group detection in  a conference paper in \cite{shi2021person}. 
In our basic approach, we used YOLO for face detection.
YOLO generated a large number of false face detections that belonged to different student groups.
To address the problem, we relied on the fact that
      student faces that are far away from the camera are characterized
      by high instantaneous-frequency components.
We thus used FM feature extraction and a simple LeNet5 network
      to remove false face detections and also detect
      back of the head students facing away from the camera.
Here, we note that the advantages of the FM representations
     come from the fact that they are explainable and provide
     additional image representations that go beyond
     the standard raw images processed by YOLO.
We will employ this system for student group detection.
     
\subsubsection{Face Recognition}
In order to recognize the student participant, following face detection,
     we use face recognition.
Here, we note that face recognition is a very mature research area
     for the case when the humans are facing the camera.
Unfortunately, this is not the case here.
We are faced with several challenges since the students
     are not  posing for the camera.
Instead, they can be at arbitrary angles.
Our approach was to adopt a state of the art system face recognition system
    and retrain it for video face recognition for our current problem.
Thus, for our baseline system, we use the InsightFace system \cite{deng2019arcface} 
    that is based on  Additive Angular Margin Loss for Deep Face Recognition  (ArcFace).
Here, we note that ArcFace has been tested on a large number of camera-facing image datasets and a variety of loss function models.
We have summarized our modified system in a conference paper in \cite{tran2021facial}.
For completeness, we will provide a summary of our methodology adopted from \cite{tran2021facial} 
     in our methods section.
     
\subsubsection{Tracking Under Occlusion}
Following person recognition and face recognition, we are faced
    with the problem of tracking under occlusion.
As mentioned in the introduction, 
    previously considered methods include
    the use of correlation filters in \cite{chen2020augmented},
    a classifier approach \cite{dong2016occlusion}, 
    convolutional neural networks in \cite{yuan2020scale},
    and  a geometric approach  in 
        \cite{nasseri2021simple} and \cite{stadler2021improving}.
As noted earlier, we will be comparing our approach
     to  the  Simple Online and Real-time Tracking with Occlusion Handling (SORT\_OH 
         \cite{nasseri2021simple})
        which achieved state-of-the-art results on the 
        MOT16/17 datasets for pedestrian tracking. 
        
We also provide a summary of other research in this area.
In  \cite{lan2021robust}, the authors present 
     a novel approach for visual object tracking that discriminates occlusion from the 
     self-deformation of the target.
In \cite{kuipers2020hard}, the authors evaluate the performance of visual object trackers
    in challenging occluded scenarios by creating a small dataset that includes sequences 
    with multiple instances of hard occlusions.
In \cite{stadler2021improving}, the authors developed a 
    regression-based multi-pedestrian tracker that can re-track targets without an extra 
    re-identification model. 
The paper reports a method for improving track management by 
    regressing inactive tracks and also developing a method for dealing
    with tracks that are out of the camera’s view.
In \cite{yuan2020scale}, the authors develop an object-tracking method 
    based on the combination of correlation filters and ResNet features. 
The paper describes the use of  response maps by extracting features 
   from different layers of ResNet, and 
   then fusing response maps using the AdaBoost algorithm. 
In \cite{chen2020augmented}, the authors propose the Kernelized 
   Correlation Filter (KCF) model to track ships in consecutive maritime images and 
   then use the tracking to estimate ship trajectories. 
In \cite{dong2016occlusion}, the authors present an integrated Circulant 
   Structure Kernels (ICSK) tracking framework to handle occlusion by estimating target
   objects’ translation and scale variations.

The paper describes a new method to support dynamic participant tracking 
    that can deal with long-term occlusions and persons entering and leaving
    the scene.
Our DPT uses a finite-state machine to track each person.
Transitions between states are based on intuitive geometrical
    constraints.
As we discuss in the results, the DPT is proven to be very effective
   on real-life AOLME videos.
       
%Papers \cite{shi2016human}\cite{shi2018dynamic}\cite{shi2018robust} and \cite{shi2021person} have provided the related human and group detectors.

%%%%%%%%%%%%%%%%%%%%%%%%%%%%%%%%%
\section{Methodology}\label{sec:methods}
\begin{figure*}[!h]
	\centering
	\includegraphics[width=\textwidth]{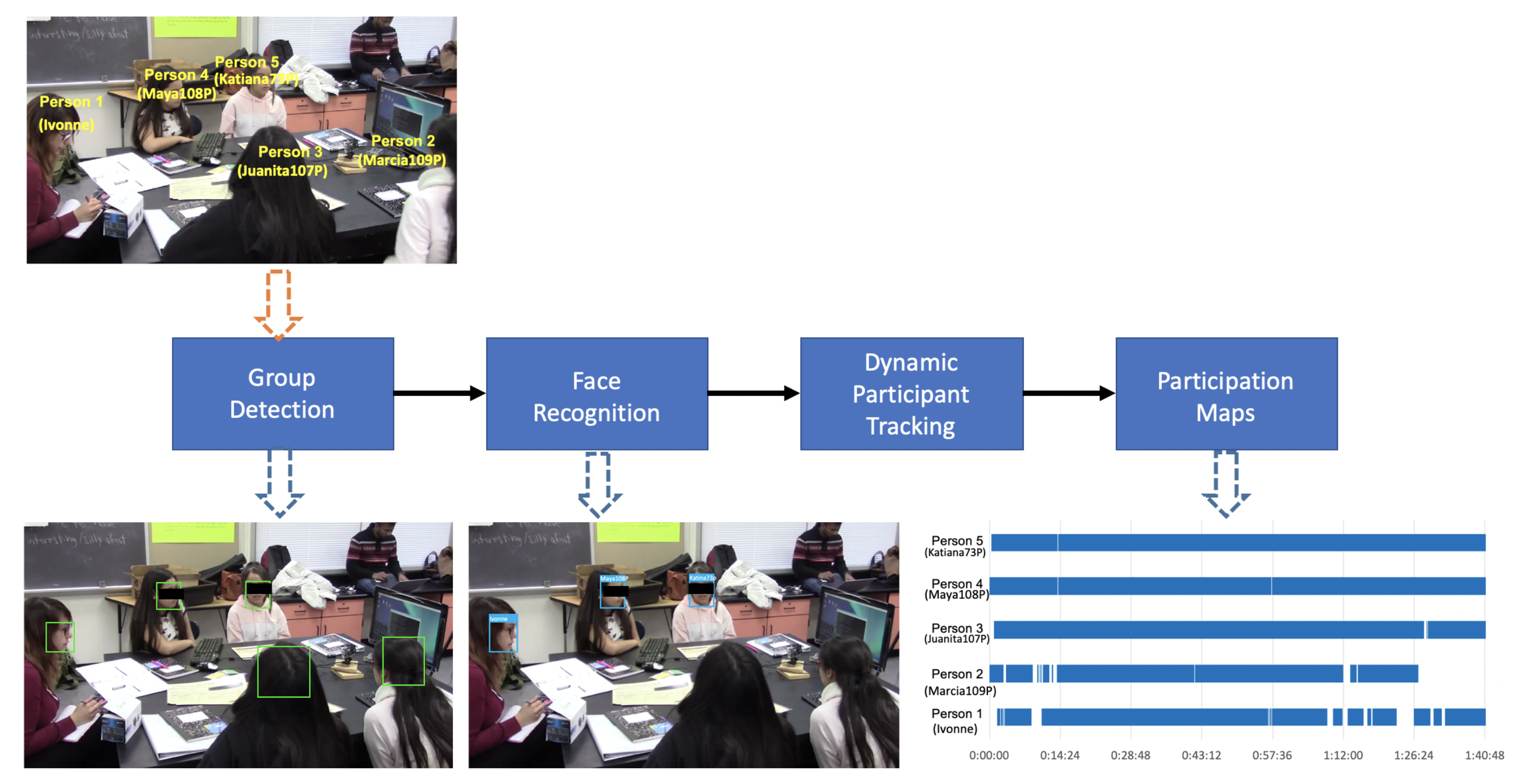}
	\caption{AOLME student participation analysis system. We detect groups every second. We perform face recognition and dynamic participant tracking every frame.} 
	\label{fig:top_level_diagram}
\end{figure*}

\subsection{Overview}
We present a top-level diagram of the entire system in Fig. \ref{fig:top_level_diagram}.
The raw input video is first processed through group detection to
        identify the students with the current group while rejecting 
        people in the background that do not belong to the current group.
We then identify the students for whom we can detect faces based 
        on a face recognition system. 
We use dynamic participant tracking for all identified students to 
      account for cases where students may move or leave the scene.
Then, we combine the information to produce participation maps documenting 
     student  participation through time.
Informed consent was obtained for all study participants.

\subsection{Group Detection}
\begin{figure*}[!h]
	\centering
	\includegraphics[width=1\textwidth]{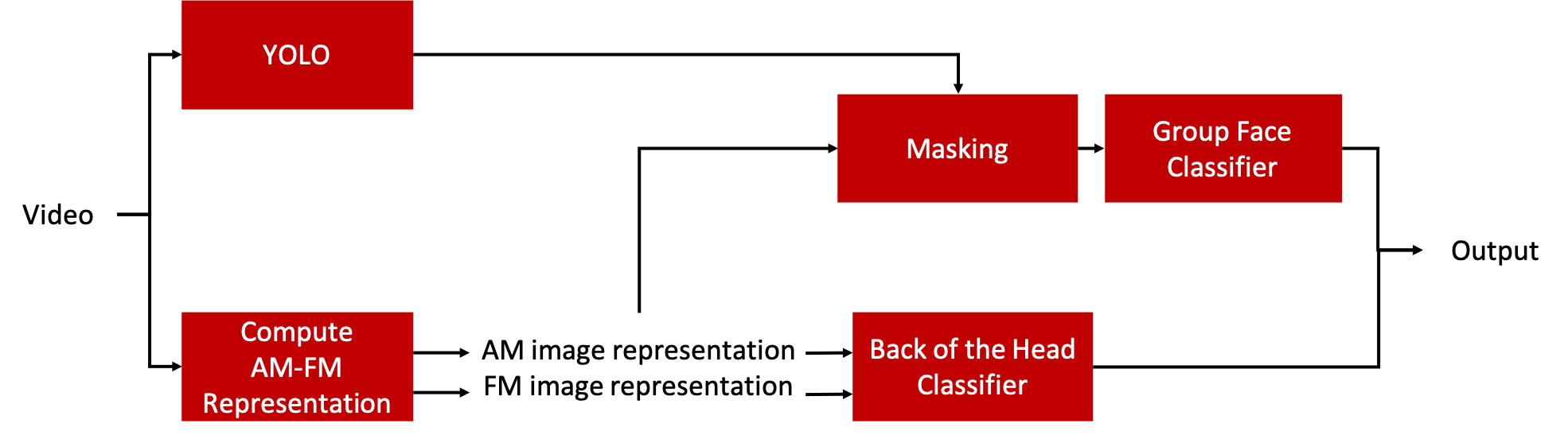}
	\caption{Student Group Detection System.} 
	\label{fig:HeadDetSystem}
\end{figure*}

For group detection, we need to detect the students sitting at the 
       table closest to the camera. 
Group detection is based on face detection for students facing the camera 
      and back-of-the-head detection for students facing away.
We present a system diagram of the group detection system in Fig \ref{fig:HeadDetSystem}. 
Due to the need for speed, we use YOLO for face detection.
Back-of-the-head detection is performed based on extracted AM-FM features as described next. 

\begin{figure*}[h!]
    \centering
    \subfigure[Classroom image.]
    {
        \includegraphics[width=0.31\textwidth]{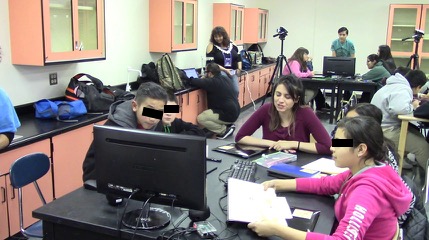}
            }
    \subfigure[AM component.]
    {
        \includegraphics[width=0.31\textwidth]{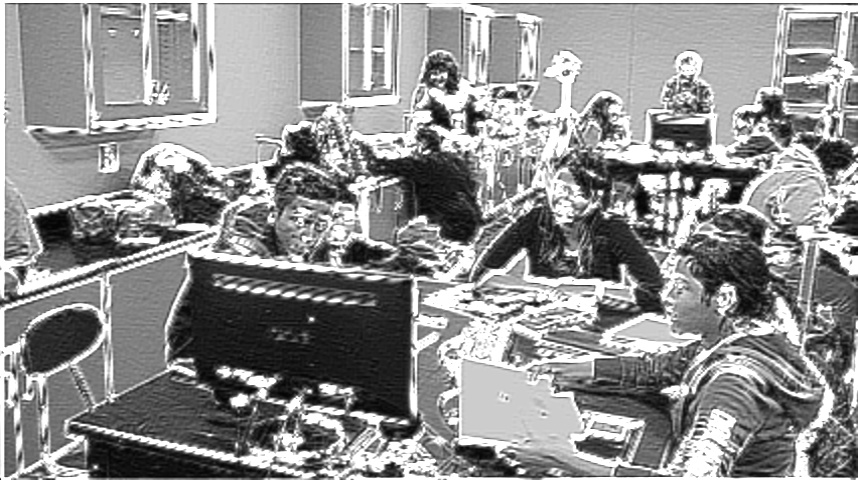}
    }
       \subfigure[FM component.]
    {
        \includegraphics[width=0.31\textwidth]{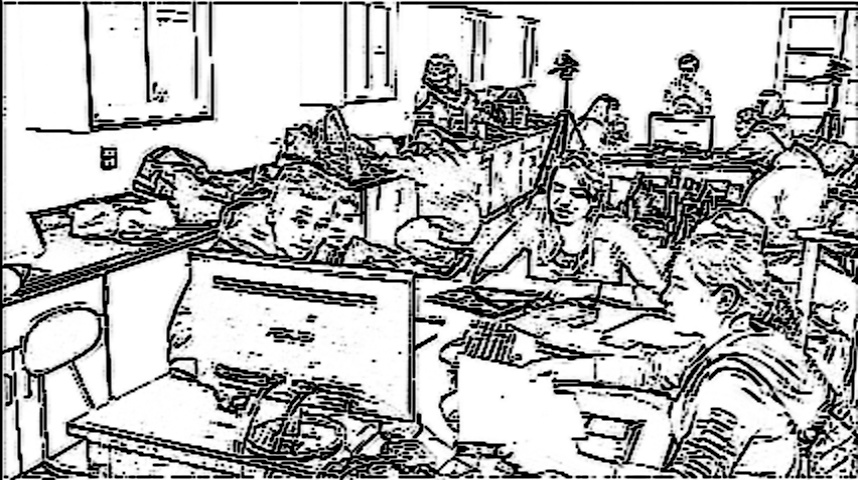}
    } 
\caption{AM-FM representation of the classroom environment.} 
\label{fig:AM-FM samples}
\end{figure*}

AM-FM components are extracted from the grayscale (Y-component) using  dominant component analysis (DCA) estimated using a 54-channel Gabor filterbank as described in \cite{shi2018robust}.
Using DCA, the input image frame is approximated by:
$ I(x, y) \approx a(x, y) \cos \varphi (x, y)$
where   $ a(x, y)$ denotes the AM component and $ \cos \varphi (x, y)$ denotes the FM component.
Fig \ref{fig:AM-FM samples} shows an example of the extracted AM-FM components.

The FM image is masked by the results of the YOLO face detector. 
We apply this step to extract the FM components over students within the desired group while rejecting detections from other groups.
FM components over the faces of the closest group exhibit lower frequency components than the higher frequency components associated with distant faces from other groups. 
To detect the group faces, we thus apply a simple, LeNet-based classifier \cite{lecun1998gradient} on the extracted FM components over $100\times100$ pixel regions.

The AM-FM components are also used to detect the hair and back-of-the-head candidate regions described in \cite{shi2018robust}. A LeNet-based classifier is used to detect the back-of-the-heads against background detections, as detailed in \cite{shi2018robust}. We detect the entire group for each video frame by concatenating the results from the face and back-of-the-head classifiers.

\subsection{Face Recognition}
We adopt the face recognition method previously described as a conference paper in \cite{tran2021facial}. We use the InsightFace \cite{guo2021sample} system to recognize faces.
The face recognition system requires a set of face prototypes associated with each participant. 

We combine sparse sampling and K-means clustering to compute face prototypes as given in Algorithm \ref{alg:kmeans} (also see \cite{tran2021facial}).

\begin{figure*}[!h]
	\centering
	\includegraphics[width=1\textwidth]{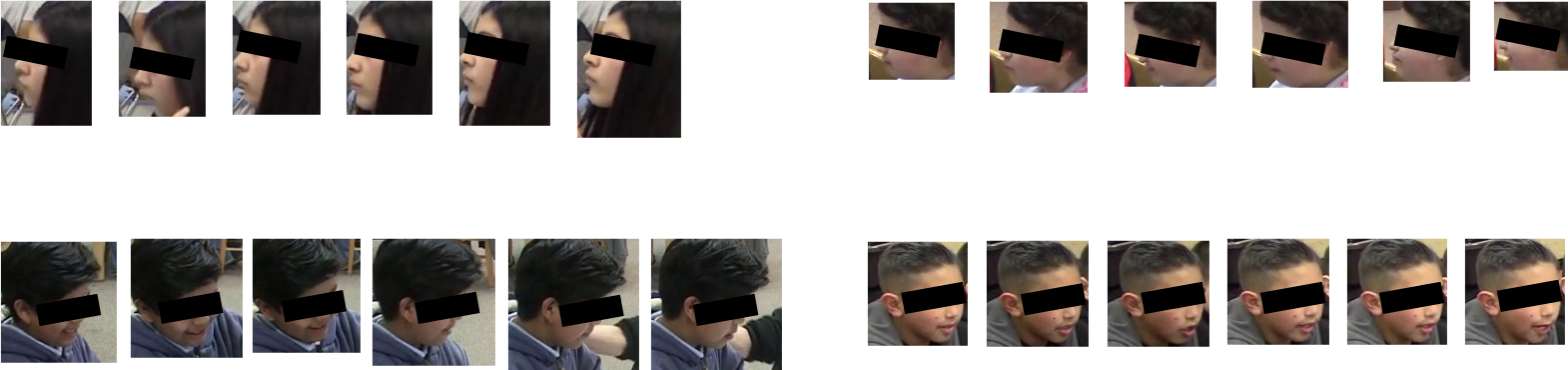}
		\caption{Face prototype samples of four students.} 
	\label{fig:face_prototypes}
\end{figure*}

Fig \ref{fig:face_prototypes} displays some samples of face prototypes.

 \begin{algorithm}[H]
 \caption{\label{alg:kmeans}Compute face prototypes using sparse sampling and K-means.}
 \begin{algorithmic}[1]
 \renewcommand{\algorithmicrequire}{\textbf{Input:}}
 \renewcommand{\algorithmicensure}{\textbf{Output:}}
 \REQUIRE Video clips associated with each participant
 \ENSURE  facePrototypes associated with each participant \\

\hspace{-0.7cm} 1: \textbf{for} each participant \\
\hspace{-0.7cm} 2: \hspace{0.5cm} \textbf{Sample} an image every 30 frames of video \\
\hspace{-0.7cm} 3: \hspace{0.5cm} \textbf{Apply} K-means clustering \\
\hspace{-0.7cm} 4: \hspace{0.5cm}  \textbf{Select} cluster means \\
\hspace{-0.7cm} 5: \hspace{0.5cm}  \textbf{Find} the nearest images from cluster centroids \\
\hspace{-0.7cm} 6: \hspace{0.5cm}  \textbf{Align} faces to $112\times112$

\end{algorithmic}
\end{algorithm}

%%%%%%%%%%%%%%%%%%%%%%%%%%
%%%%%%%%%%%%%%%%%%%%%%%%%%

%\begin{figure}[!h]
%\begin{algorithmic}[1] % The number tells where the line numbering should start
%\onehalfspacing
%\Function{Face\_Prototypes\_Generation}{}
%\inp
%\commvar Video clips associated with each participant
%\outp
%\commvar facePrototypes associated with each participant \\
%\textbf{for} each participant \\
%\hspace{1cm}  \textbf{Sample} an image every 30 frames of video \\
%\hspace{1cm}  \textbf{Apply} K-means clustering \\	
%\hspace{1cm}  \textbf{Select} cluster means \\
%\hspace{1cm}  \textbf{Find} the nearest images from cluster centroids \\
%\hspace{1cm}  \textbf{Align} faces to $112\times112$
%%$\bf{end}$  $\bf{function}$
%\EndFunction
%%\caption{\label{alg:alg_DSC} Dynamic Scene Classifier Algorithm}
%\end{algorithmic}
%\caption{\label{alg:kmeans}Compute face prototypes using sparse sampling and K-means.}
%%\singlespacing
%\end{figure}
%%%%%%%%%%%%%%%%%%%%%%%%%%
%%%%%%%%%%%%%%%%%%%%%%%%%%

\subsubsection{Sparse sampling}
To achieve sparsity, the algorithm extracts a single sample image per second of video with a frame rate equal to 30 fps.

\subsubsection{K-means clustering}
We use clustering to describe different face poses. 
The algorithm searches for the training image that 
    is closest to a cluster centroid to prevent the usage of centroids that might be impractical.
Once the algorithm has identified a prototype image that is closest to the mean, it proceeds to align and resize every image to $112\times112$ pixels. This paper uses K-means with 64 clusters for data training.

Once the face prototypes are obtained, we use the InsightFace system to identify
    the students.
Seven-fold data augmentation is implemented to increase the training dataset.
Data augmentation is based on random rescaling, cropping, rotating, and left-to-right flipping. This process generates a total of 18,816 prototype faces for training and validation.

The algorithm for face recognition uses the MTCNN model to detect faces in the video. 
It then calculates the minimum distances to the face prototypes to identify each participant.

\begin{figure*}[h!]
	\centering
	\includegraphics[width=\textwidth]{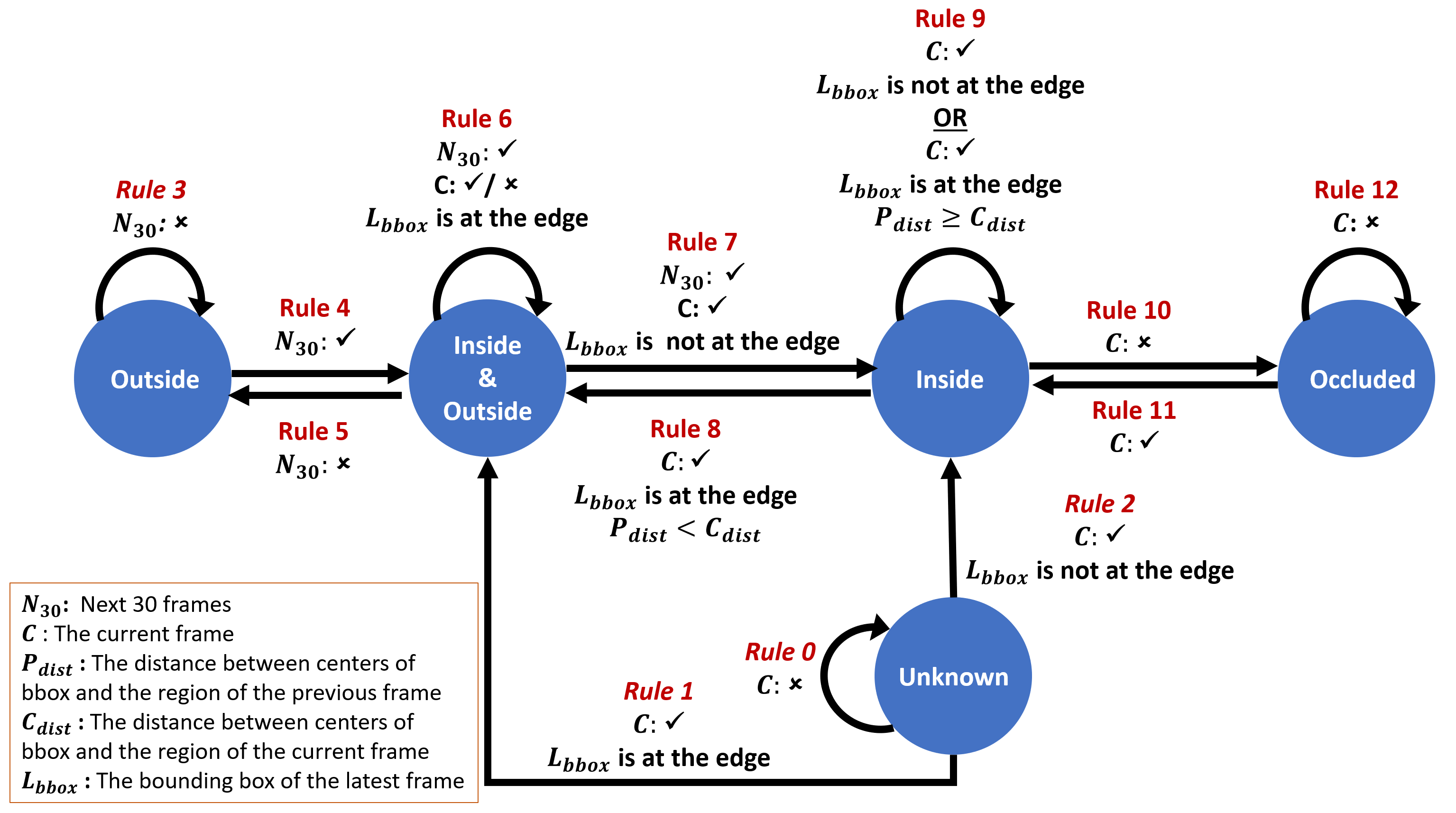}
	\caption{Dynamic participant tracking system. 
	              Here, bbox refers to the bounding box.}
	\label{fig:DSC}
\end{figure*}

\subsection{Dynamic Participant Tracking}
We develop the Dynamic Participant Tracking (DPT) system to account for the presence of the students in relation to the camera as shown in Fig.\ref{fig:DSC}. More specifically, during the tracking process, a participant is classified as being `Inside' or `Outside' the video frame, in the process of leaving the scene (`Inside \& Outside'), occluded by another object (`Occluded'), or being in an undetermined state (`Unknown'). In what follows, we begin the section by providing definitions of each state. We then describe how to determine whether a participant is in one of the states and how to transition from state to state. Here, we note that state transitions are based on the current state and the participant detection results. We also note that the DPT is applied separately for each participant.

\subsubsection{Edge}
\begin{figure}[!h]
	\centering
	\includegraphics[width=0.5\textwidth]{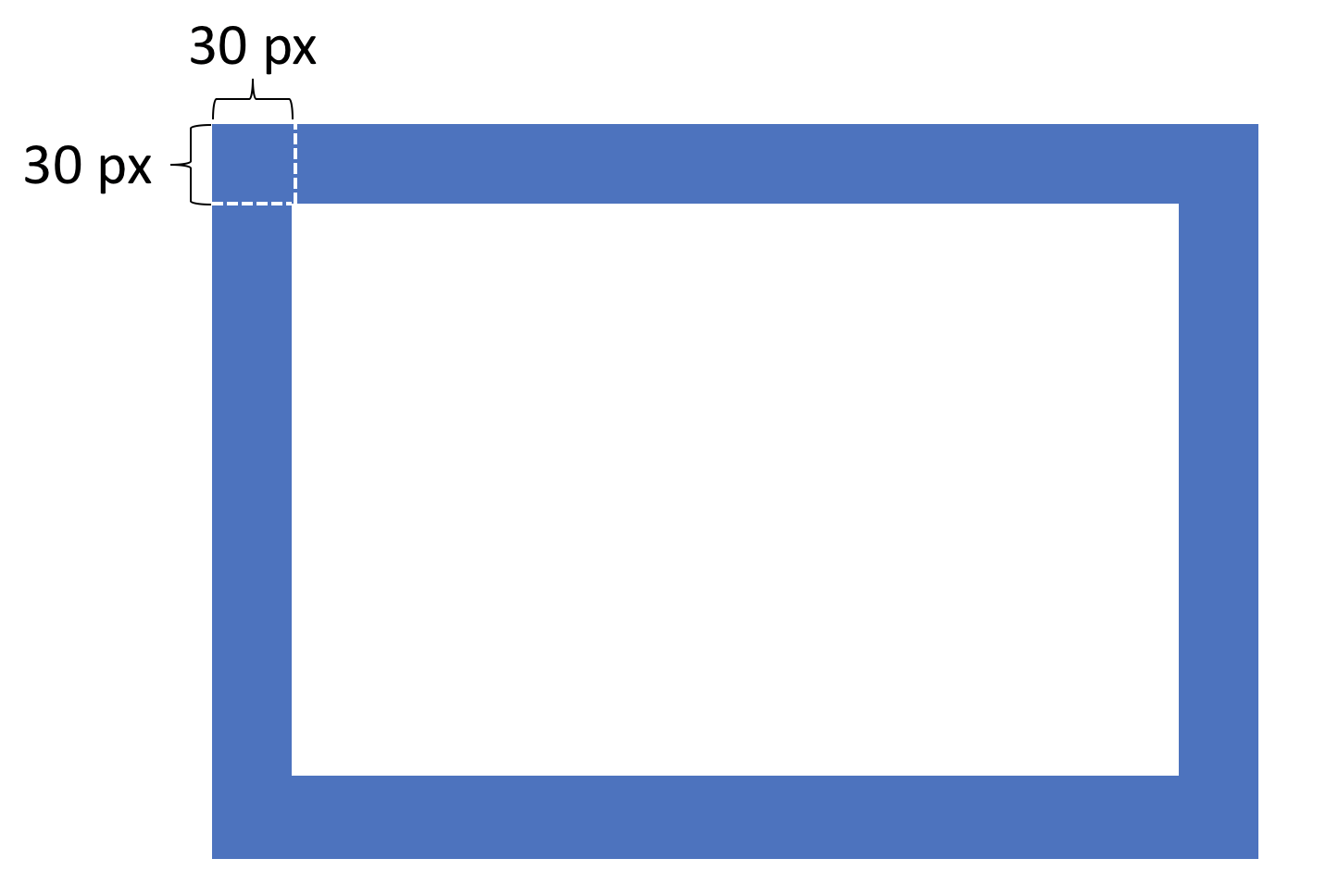}
	\caption{The edge of the video frame is defined as the set of pixels located within 30 pixels of the edge of the frame.}
	\label{fig:edge}
\end{figure}

We define the edge of the video frame to be the pixels less than 30 pixels from the edge, as shown in Fig. \ref{fig:edge}.

\subsubsection{The distance between centers}
\begin{figure}[!h]
    \centering
    \subfigure
    {
        \includegraphics[width=0.45\textwidth]{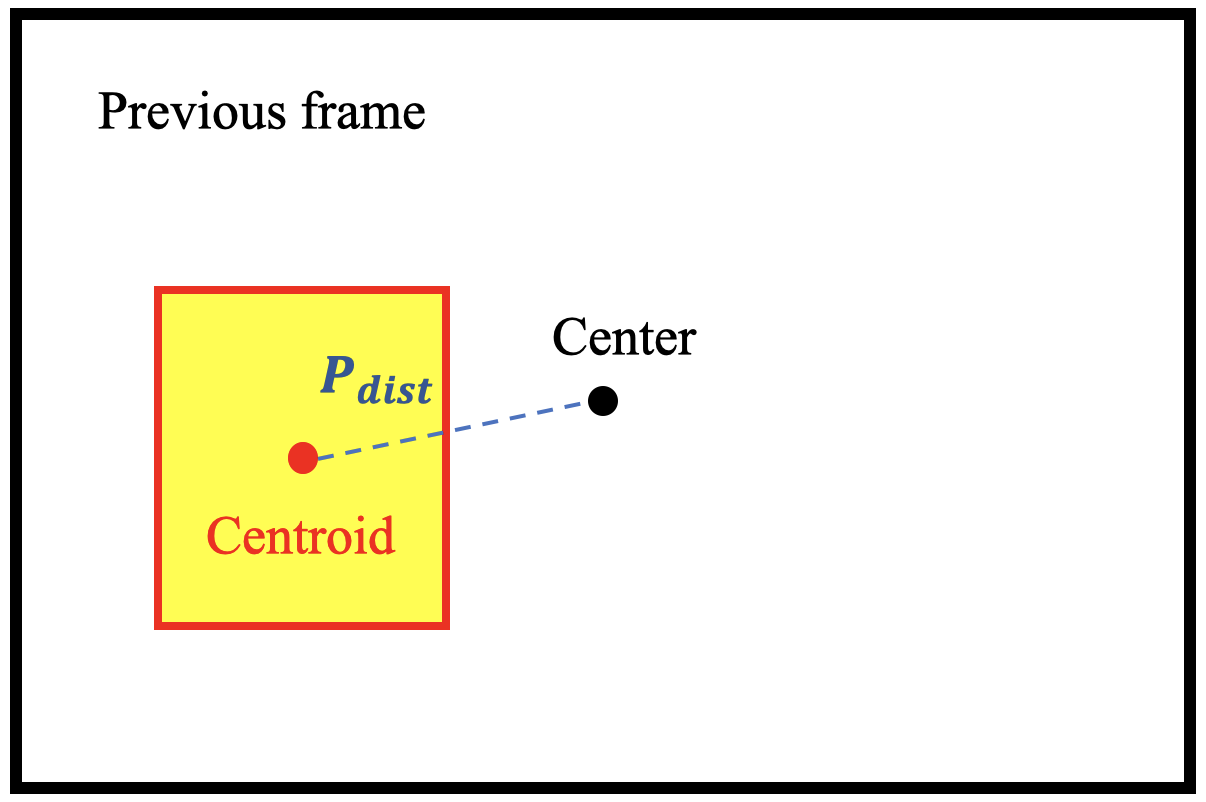}
        %\label{fig:second_sub}
    }
    \\
    \subfigure
    {
        \includegraphics[width=0.45\textwidth]{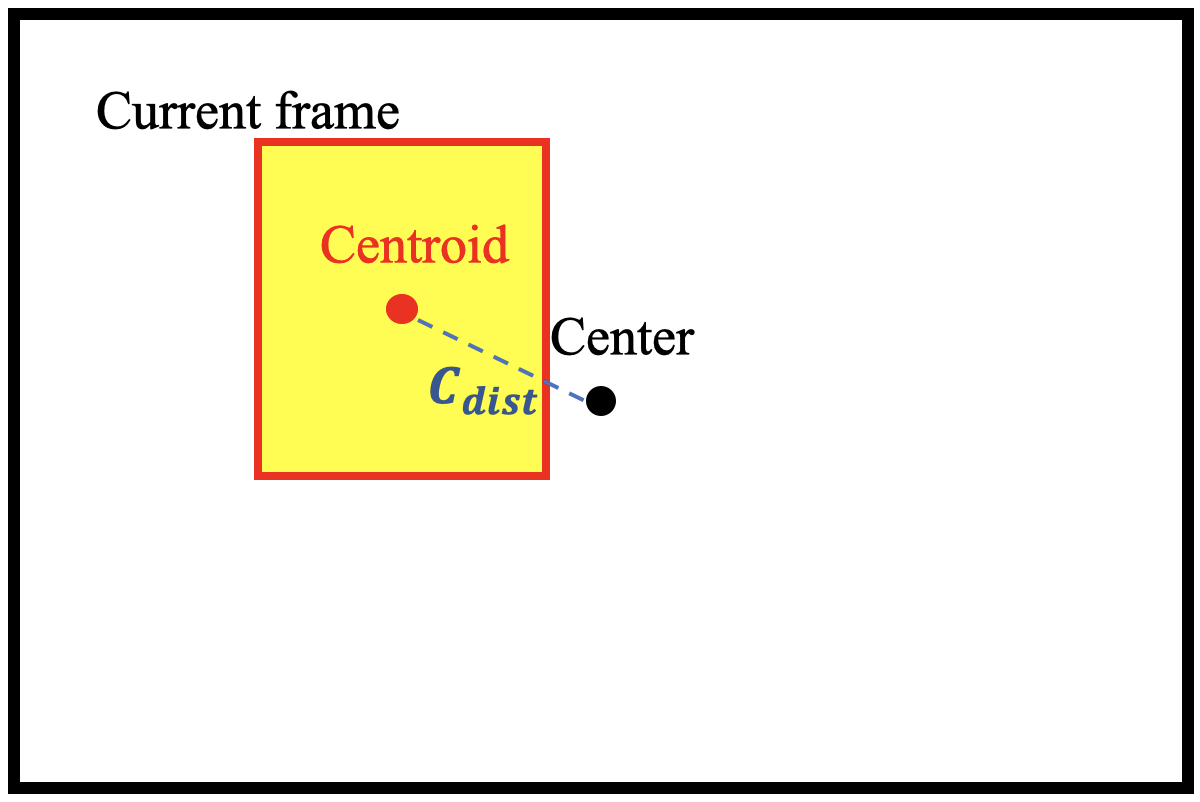}
       % \label{fig:third_sub}
    }
    \caption{State determination is based on the distance from the center of the frame to the centroid of the object detection bounding box.}
    \label{fig:centroid_dist}
\end{figure}

We define the centroid distance D (Fig. \ref{fig:centroid_dist}) between the 
      center of the frame and a bounding box that exists in the frame using: 
 $$ D = \sqrt{|x2-x1|^2+|y2-y1|^2}.    $$

\subsubsection{States}
The state of each participant is based on the location of the bounding box resulting 
      from participant detection. 
All participants are initially in `Unknown' state if not detected. 
This state remains unknown as long as they are not detected in the current state. 

If a person is detected in the initial frame, their location is used to determine their initial state. Thus, if the person is detected entirely inside the frame, their state is set as `Inside'. 
If the person is detected at the edge of the frame, their initial state is set to 
   `Inside \& Outside'.

\begin{figure}[!h]
	\centering
	\includegraphics[width=0.45
	\textwidth]{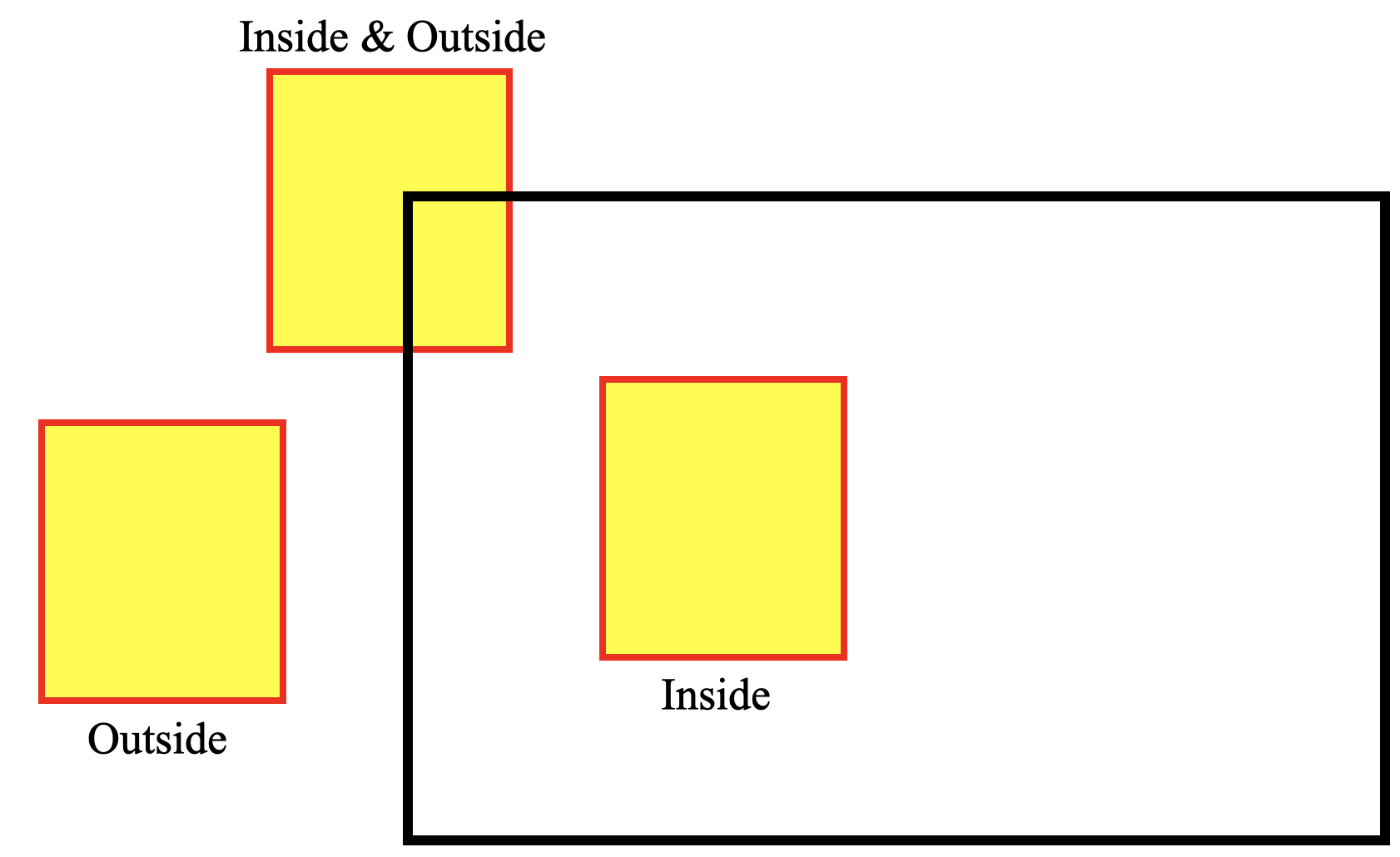}
	\caption{Definitions of `Outside', `Inside \& Outside', and `Inside' states.}
	\label{fig:out_inside_definition}
\end{figure}

%%%%%%%%%%%%%%%%%%%%%%%%%%%%%%%%%%%%
In Fig.\ref{fig:out_inside_definition},
    we demonstrate how the locations of the bounding boxes 
    are used to determine the states `Outside’, `Inside \& Outside’, and `Inside’. 
The black rectangle represents the frame edge.
The red-yellow rectangles represent the bounding boxes.
This frame has no bounding box if a person is occluded because 
      other people or objects cover them.

\subsubsection{Inputs}
We define the five possible inputs for determining transitions between states as follows. 

We use $C$ to denote the detection in the current state. Thus, $C:$ \checkmark \ denotes successful detection. On the other hand, $C:$ \xmark \ denotes failure to detect a participant in the current frame. 

We use $N_n$ to denote the detection of a participant over $n$ frames. Thus, $N_n:$ \checkmark \ denotes successful detection in any of the following $n$ frames while $N_n:$ \xmark \ denotes failure to detect a participant in the subsequent $n$ frames. 

We use $P_{dist}$ to denote the distance between the centroid of the bounding box and the center of the previous frame.

We use $C_{dist}$ to denote the distance between the bounding box's centroid and the current frame's center (See Fig.\ref{fig:centroid_dist}).

We use $L_{bbox}$ to denote the location of the latest detection with a bounding box.

%%%%%%%%%%%%%%
\subsubsection{DPT transitions}
The initial states can be `Inside', `Unknown', or `Inside \& Outside' as described in the DPT states subsection. Here, we describe transitions among other states. We note that for each state, we consider all possible input combinations for determining how to transition to another state.

\begin{figure}[!h]
	\centering
	\includegraphics[width=0.45\textwidth]{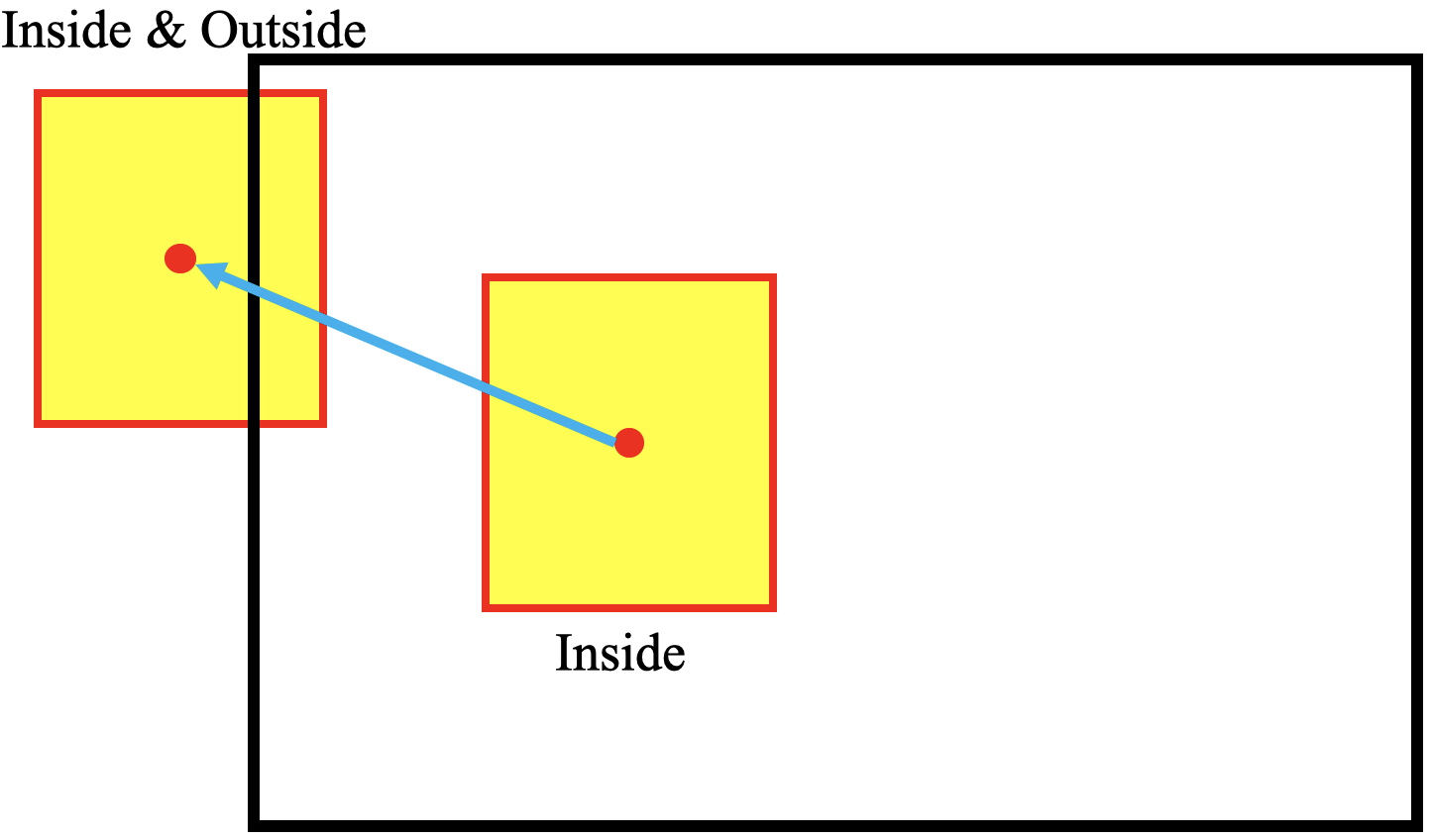}
	\caption{One example of DPT transition rules: rule 8.}
	\label{fig:rule_8}
\end{figure}

From the `Inside' state, a participant can move to the `Inside \& Outside' state, `Inside' state, or the `Occluded' state as given below:
\begin{itemize}
\item To transition to the `Inside \& Outside' state, we detect a movement inside the frame toward outside the frame. Here, the movement is detected by requiring that 1) the person is detected in the current frame, 2) the centroid's distance of previous frame is less than the one of current frame, and 3) the person is detected at the edge of the current frame. We simplify these rules by using:
 $C:$ \checkmark,  $P_{dist} < C_{dist}$, and $L_{bbox}$ is at the edge (rule 8) (See Fig.\ref{fig:rule_8}).
\item To remain in the `Inside' state, we detect a movement inside the frame. Here, the movement is detected by requiring that 1) $C:$ \checkmark and $L_{bbox}$ is not at the edge, or 2) $C:$ \checkmark, $P_{dist} \ge C_{dist}$, and $L_{bbox}$ is at the edge (rule 9). 
\item To transition to the `Occluded' state, we detect the movement disappears inside the frame. Here, it requires that $C:$ \xmark \ (rule 10). 
\end{itemize}

From the `Outside' state, a participant can move to the `Outside' state or the `Inside \& Outside' state as given below:
\begin{itemize}
\item To remain in the `Outside' state, we detect no movement in the frame. Here, it requires that $N_n:$  \xmark \ (rule 3). 
\item To transition to the `Inside \& Outside' state, we detect a movement from outside the frame toward inside the frame. The movement is detected by requiring $N_n:$  \checkmark \ (rule 4). 
\end{itemize}

From the `Inside \& Outside' state, a participant can move to the `Outside' state, `Inside \& Outside' state, or the `Inside' state as given below:
\begin{itemize}
\item To transition to the `Outside' state, we detect a movement from the frame toward outside the frame. The movement is detected by requiring $N_n:$  \xmark \ (rule 5). 
\item To remain in the `Inside \& Outside' state, we detect a movement around the edge of the frame. Here, the movement is detected by requiring that $N_n:$  \checkmark \ , $C:$ \checkmark / \xmark \ , and $L_{bbox}$ is at the edge (rule 6). 
\item To transition to the `Inside' state, we detect a movement from outside the frame toward inside the frame. Here, the movement is detected by requiring that $N_n:$  \checkmark \ , $C:$ \checkmark \ , and $L_{bbox}$ is not at the edge (rule 7). 
\end{itemize}

From the `Occluded' state, a participant can move to the `Inside' state or the `Occluded' state as given below:
\begin{itemize}
\item To transition to the `Inside' state, we detect a movement that appears inside the frame. The movement is detected by requiring $C:$ \checkmark \ (rule 11). 
\item To remain in the `Occluded' state, we detect no movement inside the frame. Here, it requires that $C:$ \xmark \ (rule 12). 
\end{itemize}

From the `Unknown' state, a participant can move to the `Unknown' state, `Inside \& Outside' state, or the `Inside' state as given below:
\begin{itemize}
\item To remain in the `Unknown' state, we detect no movement inside the frame. Here, it requires that $C:$ \xmark \ (rule 0). 
\item To transition to the `Inside \& Outside' state, we detect a movement that appears in the frame. Here, the movement is detected by requiring that $C:$ \checkmark \ and  $L_{bbox}$ is at the edge (rule 1). 
\item To transition to the `Inside' state, we detect a movement that appears in the frame. Here, the movement is detected by requiring that $C:$ \checkmark \ and  $L_{bbox}$ is not at the edge (rule 2). 
\end{itemize}

In this paper, for $N_n$, we set up $n=30$ because the frame rate of video is 30 fps and students in the videos have big movement.
We note that $n=30$ represents a second.
Here, it is important to note that our parameters were intuitively set for 1-second transitions.

%$\bf{greatest} $

\begin{figure}[!b]
    \centering
     \subfigure
 {
        \includegraphics[width=0.45\textwidth]{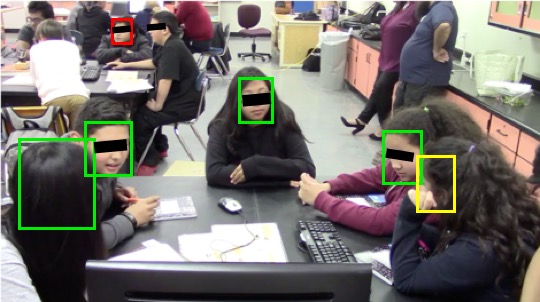}
        %\label{fig:second_sub}
    }\\
    \subfigure
    {
        \includegraphics[width=0.45\textwidth]{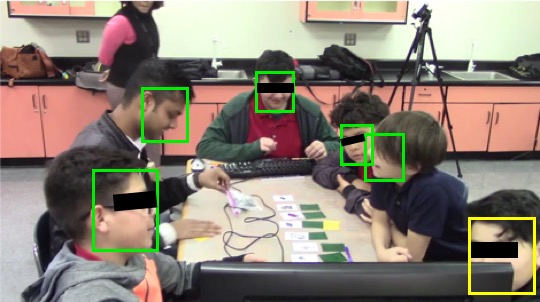}
        %\label{fig:second_sub}
    }
 \caption{Head detection system results. 
True positives are bounded by green boxes.
False positives are bounded by red boxes.
False negatives are bounded by yellow boxes.
For successful detection, we require the intersection over union (IOU) 
      score to be at least 0.6.} 
\label{fig:successful and failed group det cases}
\end{figure}

\begin{figure*}[t!]
    \centering
    \subfigure[YOLO: Example 1]
    {
        \includegraphics[width=0.31\textwidth]{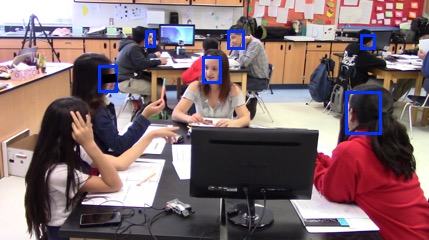}
            }
    \subfigure[Ground Truth: Example 1]
    {
        \includegraphics[width=0.31\textwidth]{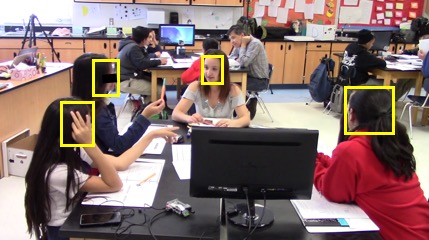}
    }
       \subfigure[Ours: Example 1]
    {
        \includegraphics[width=0.31\textwidth]{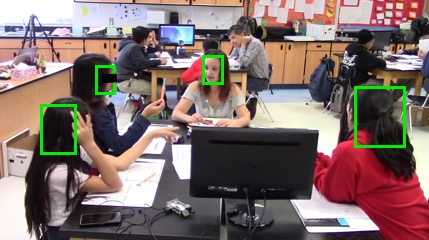}
    } \\
     \subfigure[YOLO: Example 2]
         {
        \includegraphics[width=0.31\textwidth]{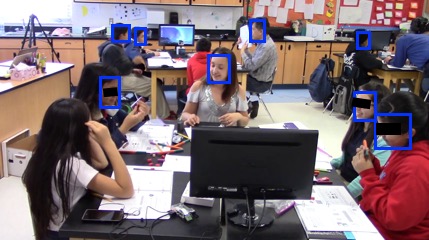}
    }
    \subfigure[Ground Truth: Example 2]
    {
        \includegraphics[width=0.31\textwidth]{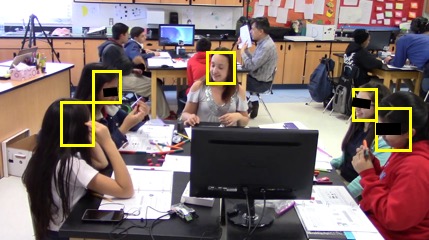}
    }
       \subfigure[Ours: Example 2]
    {
        \includegraphics[width=0.31\textwidth]{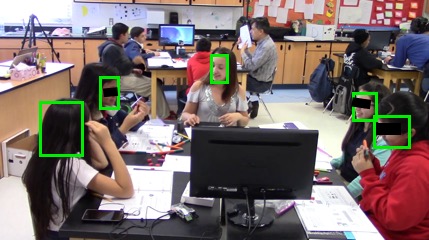}
    }    \caption{Examples from Group Detection Results. 
Left column   ((a) and (d)) shows the results of YOLO.
Middle-column ((b) and (e)) shows the ground truth.
Right-column  ((c) and (f)) shows the results of the proposed method.} 
\label{fig:yolo_gt_comb_comparison}
\end{figure*}

\section{Results}\label{sec:results}
We summarize our results over the final testing datasets
     (see Table \ref{table:aolme_student_datasets}).
First, we summarize our group detection results over
     the massive AOLME-GT dataset.
For group detection, recall that we labeled 12,518,250 student instances in over
      21 hours and 22 minutes of real-life videos (see section \ref{sec:AOLME-GT}).     
For system testing, we will first present results over the 
      carefully selected short videos of the AOLME-DST dataset 
      (see section \ref{sub:AOLME-DST}).
We also summarize final system testing results over raw, real-life
      videos of the AOLME-DLT dataset
      (see section \ref{sec:AOLME-DLT}).
We then present participation maps for visualizing the final results.
As mentioned earlier, the testing datasets do not share any
      video sessions with the training and validation datasets.
      
\subsection{Student group detection testing using AOLME-GT dataset}      
We begin with a simple example in Fig. \ref{fig:successful and failed group det cases}.
For students within the group of interest, we use green
     bounding boxes to indicate successful detections (true positives (TP)).
We used yellow bounding boxes to denote false negatives (FN), when
     we fail to detect a student that belongs to the group.
For students outside the group, we use red bounding
     boxes to indicate false positives (FP).
In the top image of Fig. \ref{fig:successful and failed group det cases},
     we see that we have a false positive case for a 
     background student facing the camera.
Then, in the bottom image of Fig. \ref{fig:successful and failed group det cases},
     we have a false negative example where we could not detect
     a student that is partially occluded by the camera.
Here, it is important to note that the false negative case can 
     be corrected using the dynamic participant tracking algorithm.
Assuming that a student was detected in an earlier frame,
      the DPT will correctly classify the student as occluded and mark
      them as present.                    
         
We present comparative results of the proposed method and YOLO
      in Fig. \ref{fig:yolo_gt_comb_comparison}.
To differentiate among methods, we use
      blue bounding boxes for YOLO (left column),
      yellow bounding boxes for the ground truth (middle column), and
      green  bounding boxes for the proposed method (right column).      
In this example, the proposed method successfully detected the entire
      group without giving any false positives.
In contrast, YOLO, trained on the same dataset,
      gave several false positives by wrongly labeling
      background face detections as being part of the group.
Most alarmingly, YOLO failed to detect a student that belongs to the group
      (see student in the lower-left part of the image in the left column).          
Our performance clearly benefited from the use of multiple representations
     and our back-of-the-head detector.
     
We provide comparative results over the AOLME-GT dataset 
       in Table \ref{table:long_video_group_det_results}.
Here, note that the numbers of detected persons are 
      often higher than the number of 
       labels as both methods may falsely identify background students
       as belonging to the current student group.
Furthermore, false positives are associated with falsely labeling
      out of group students as being part of the group.
On the other hand, false negatives are primarily due to occlusions.
Our approach achieves a substantially lower number of false positives.             
We use the F1 score to assess overall performance (harmonic mean of precision
      and recall).
We note that our proposed method performs better on all video examples
      (except for V9 where performance was the same).      
In many cases, the proposed method is substantially better, with
      over 0.07 improvement
      (e.g., V1 improved by 0.09, 
               V2 improved by 0.14, 
               V4 \& V6 improved by 0.07). 
Overall,  it yields an F1 score of 0.85 against 0.80 for YOLO.       

% Please add the following required packages to your document preamble:
% \usepackage{multirow}
\begin{table*}[!t]
\caption{\label{table:long_video_group_det_results}
 Comparative results for student group detection over 13 videos.
 TP, FP, and FN refer to true positives, false positives, and false negatives,
     respectively.
 F1 scores are given for each video and each method.    
 The videos represent different student groups based on the AOLME-GT dataset.
 Here, recall that there is a large number of students in each image that
       do not belong to the current group.
 Hence, the number of detected persons is higher than the number of labeled students.
}
\begin{center}
%\resizebox{\textwidth}{!}{
\begin{tabular}{lllllllll}
\toprule %
\textbf{Video}                & \textbf{Length}          & \begin{tabular}[c]{@{}l@{}}Labeled \\ Persons\end{tabular} & \textbf{Method}          & \begin{tabular}[c]{@{}l@{}}Detected \\ Persons\end{tabular} & \textbf{TP} & \textbf{FP} & \textbf{FN} & \textbf{F1}   \\ \midrule
\multirow{2}{*}{\textbf{V1}}  & \multirow{2}{*}{96 min}  & \multirow{2}{*}{1,627,320} & \textbf{YOLO}            & 1,915,935                 & 1,153,959   & 761,976     & 124,527     & 0.72          \\
                              &                          &                            & \textbf{Proposed Method} & 1,397,790                 & 1,183,630   & 214,160     & 344,640     & \textbf{0.81} \\  \midrule
\multirow{2}{*}{\textbf{V2}}  & \multirow{2}{*}{85 min}  & \multirow{2}{*}{887,700}   & \textbf{YOLO}            & 1,274,429                 & 723,283     & 551,146     & 12,153      & 0.72          \\
                              &                          &                            & \textbf{Proposed Method} & 847,250                   & 728,110     & 119,140     & 110,140     & \textbf{0.86} \\  \midrule
\multirow{2}{*}{\textbf{V3}}  & \multirow{2}{*}{117 min} & \multirow{2}{*}{1,063,300} & \textbf{YOLO}            & 792,291                   & 720,762     & 71,529      & 321,159     & 0.79          \\
                              &                          &                            & \textbf{Proposed Method} & 819,700                   & 745,880     & 73,820      & 293,640     & \textbf{0.80} \\  \midrule
\multirow{2}{*}{\textbf{V4}}  & \multirow{2}{*}{108 min} & \multirow{2}{*}{1,139,850} & \textbf{YOLO}            & 1,212,963                 & 839,450     & 373,513     & 120,252     & 0.77          \\
                              &                          &                            & \textbf{Proposed Method} & 950,210                   & 859,290     & 90,920      & 242,850     & \textbf{0.84} \\  \midrule
\multirow{2}{*}{\textbf{V5}}  & \multirow{2}{*}{88 min}  & \multirow{2}{*}{1,233,540} & \textbf{YOLO}            & 1,380,190                 & 1,047,410   & 266,980     & 84,780      & 0.86          \\
                              &                          &                            & \textbf{Proposed Method} & 1,046,000                 & 1,008,042   & 4,345        & 205,507     & \textbf{0.91} \\  \midrule
\multirow{2}{*}{\textbf{V6}}  & \multirow{2}{*}{103 min} & \multirow{2}{*}{1,162,740} & \textbf{YOLO}            & 1,988,410                 & 1,274,280   & 591,570     & 110,490     & 0.78          \\
                              &                          &                            & \textbf{Proposed Method} & 1,488,619                 & 1,166,396   & 120,746     & 294,325     & \textbf{0.85} \\  \midrule
\multirow{2}{*}{\textbf{V7}}  & \multirow{2}{*}{90 min}  & \multirow{2}{*}{667,500}   & \textbf{YOLO}            & 955,580                   & 503,000     & 407,970     & 26,900      & 0.70          \\
                              &                          &                            & \textbf{Proposed Method} & 832,695                   & 465,002     & 291,258     & 41,159      & \textbf{0.74} \\  \midrule
\multirow{2}{*}{\textbf{V8}}  & \multirow{2}{*}{111 min} & \multirow{2}{*}{915,370}   & \textbf{YOLO}            & 967,110                   & 825,200     & 116,530     & 58,950      & 0.90          \\
                              &                          &                            & \textbf{Proposed Method} & 932,898                   & 823,727     & 73,928      & 54,368      & \textbf{0.93} \\  \midrule
\multirow{2}{*}{\textbf{V9}}  & \multirow{2}{*}{108 min} & \multirow{2}{*}{946,710}   & \textbf{YOLO}            & 1,112,350                 & 810,120     & 250,720     & 48,510      & \textbf{0.84} \\
                              &                          &                            & \textbf{Proposed Method} & 1,085,975                 & 783,624     & 235,303     & 52,991      & \textbf{0.84} \\  \midrule
\multirow{2}{*}{\textbf{V10}} & \multirow{2}{*}{106 min} & \multirow{2}{*}{712,050}   & \textbf{YOLO}            & 669,090                   & 657,890     & 6,740       & 50,290      & 0.96          \\
                              &                          &                            & \textbf{Proposed Method} & 684,413                   & 655,464     & 6,497       & 37,016      & \textbf{0.97} \\  \midrule
\multirow{2}{*}{\textbf{V11}} & \multirow{2}{*}{83 min}  & \multirow{2}{*}{797,580}   & \textbf{YOLO}            & 1,002,010                 & 677,390     & 285,170     & 32,510      & 0.81          \\
                              &                          &                            & \textbf{Proposed Method} & 950,187                   & 660,054     & 239,776     & 34,892      & \textbf{0.83} \\  \midrule
\multirow{2}{*}{\textbf{V12}} & \multirow{2}{*}{106 min} & \multirow{2}{*}{829,930}   & \textbf{YOLO}            & 848,310                   & 615,790     & 159,280     & 122,030     & 0.81          \\
                              &                          &                            & \textbf{Proposed Method} & 615,088                   & 584,524     & 3,158       & 236,436     & \textbf{0.83} \\  \midrule
\multirow{2}{*}{\textbf{V13}} & \multirow{2}{*}{81 min}  & \multirow{2}{*}{534,660}   & \textbf{YOLO}            & 528,700                   & 465,340     & 40,080      & 46,100      & 0.92          \\
                              &                          &                            & \textbf{Proposed Method} & 505,876                   & 453,333     & 17,181      & 49,941      & \textbf{0.93} \\ \midrule
\multirow{2}{*}{\textbf{Total}} & \multirow{2}{*}{21h 22m}  & \multirow{2}{*}{12,518,250}   & \textbf{YOLO}            & 14,647,368                   & 10,313,874     & 3,883,204      & 1,158,651      & 0.80          \\
                              &                          &                            & \textbf{Proposed Method} & 12,156,701                   & 10,117,076     & 1,490,232      & 1,997,905      & \textbf{0.85} \\ \bottomrule
\end{tabular}
\end{center}
\end{table*}

\subsection{Dynamic Participant Tracking and final system testing results}
This section provides comparative results of the DPT against SORT\_OH
     as well as results over the raw, real-life video sessions.
We also present the use of participation maps for visualizing 
     the final results.     

Following student group detection, we
      compare the performance of the DPT (proposed method)
       against SORT\_OH for the AOLME-DST test dataset
      (see section \ref{sub:AOLME-DST}).
Our results include detailed performance analysis 
      for each student participant in Tables
      \ref{table:occ_det_g1}, \ref{table:occ_det_g2},
      \ref{table:occ_det_g3}, \ref{table:occ_det_g4}, and
      \ref{table:occ_det_g5}.
      
We note that our proposed method performs exceptionally on
      nearly every case.
In the overwhelming majority of the test cases,
     we are able to dynamically track each participant with 100\% accuracy.   
On the other hand, SORT\_OH fails to track several students.
Here, we define failure as the inability of the method to track students
     with more than 70\% accuracy.
We highlight failure examples in red in Tables
      \ref{table:occ_det_g1}, \ref{table:occ_det_g2},
      \ref{table:occ_det_g3}, \ref{table:occ_det_g4}, and
      \ref{table:occ_det_g5}.
In our 35 test video sequences, we can see 14 examples of failures by SORT\_OH.
Out of the five groups, we can see that  SORT\_OH has at-least one failure to track example
     in each student group. 
In comparison, our proposed method failed on just one example (see Table \ref{table:occ_det_g4}).     

We believe that the efficacy of the DPT is due to its simplicity.
The finite-state transitions were derived based on intuitive rules that did not require training.
The only parameter used by DPT is to require the persistence of each transition
       rule over 30 frames (=1 second).

We report final testing results over raw, real-life video sessions
    of 23 minutes and 45 seconds of the AOLME-DLT dataset
      (see section \ref{sub:AOLME-DLT}).
In Table \ref{table:occ_rec}, we compare DPT against not using any tracking.
The results clearly illustrate the need for dynamic tracking.
The overall accuracy improved from 61.9\% to 82.3\% when using the DPT.

\begin{figure}[!h]
    \centering
     \subfigure[Ground truth participation map.]
 {
        \includegraphics[width=0.48\textwidth]{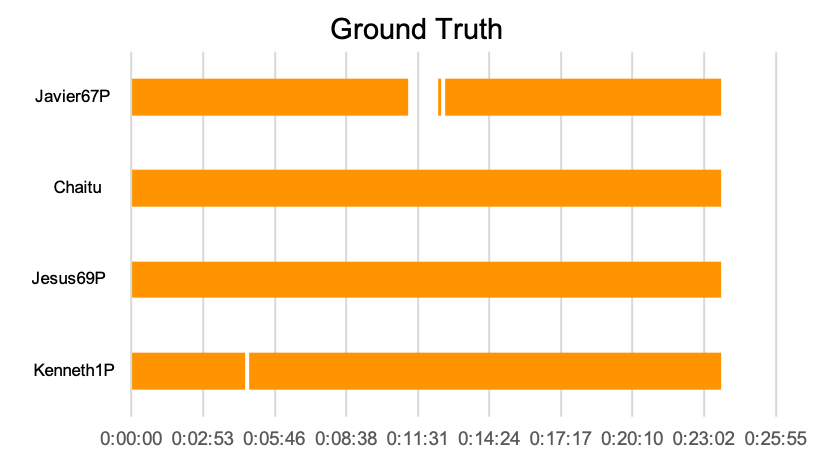}
        %\label{fig:second_sub}
    }\\
    \subfigure[Participation map without DPT. The average accuracy is 60.3\%.]
    {
        \includegraphics[width=0.48\textwidth]{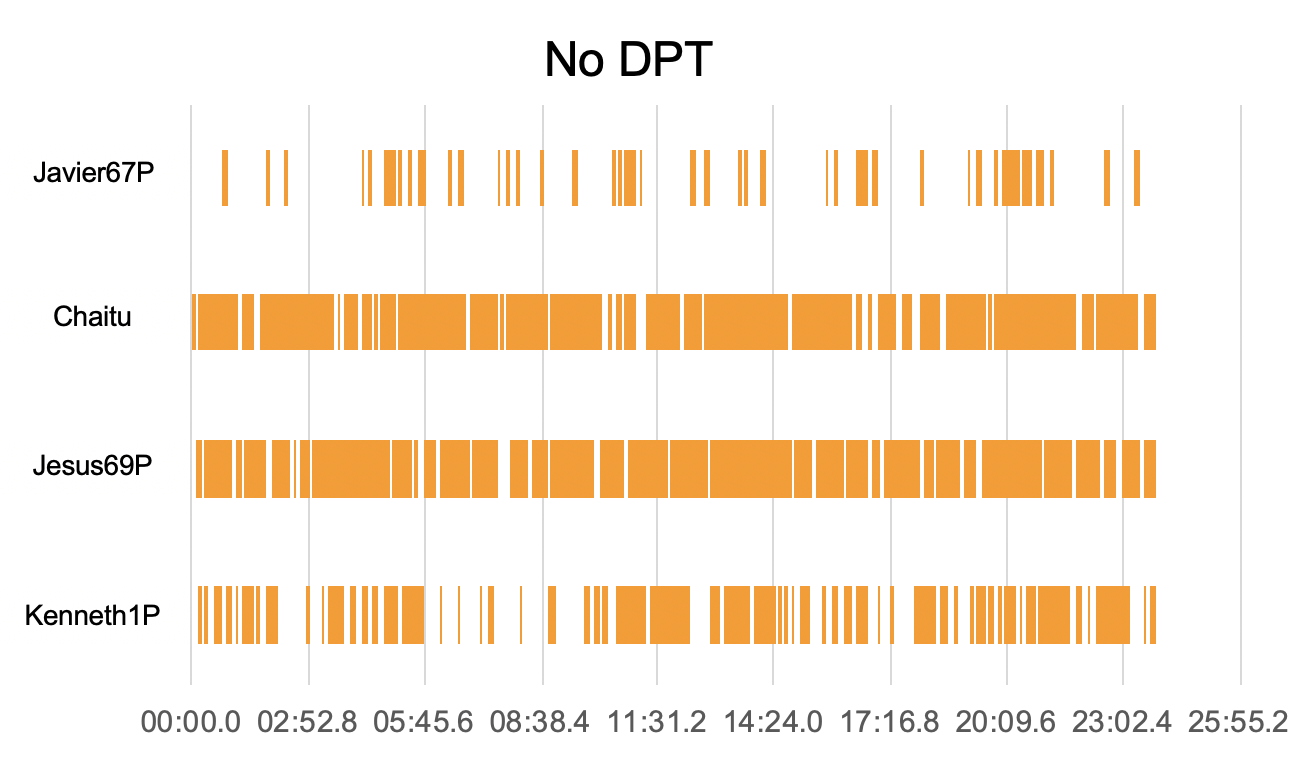}
        %\label{fig:second_sub}
    }
    \\
    \subfigure[Participation map with DPT. The average accuracy is 76.7\%.]
    {
        \includegraphics[width=0.48\textwidth]{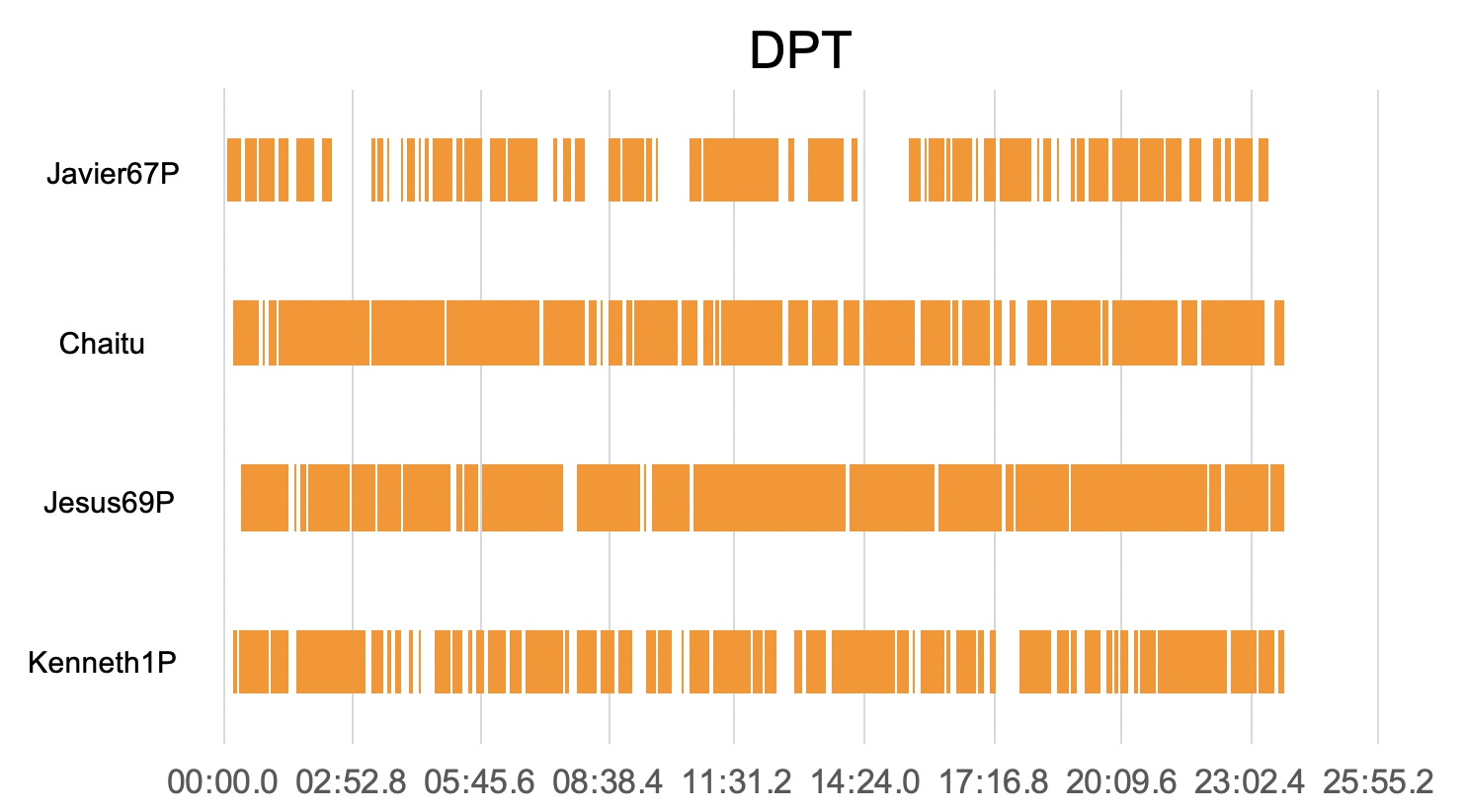}
        }

    \caption{
    Student participation maps on long videos of AOLME-DLT dataset
            (see V1 in Table \ref{table:occ_rec}).
    The results demonstrate the effectiveness of DPT (bottom)
           against ground truth (top).
    Without DPT, the lack of detection gives several false negatives (middle).
    }
    \label{fig:ActivityMaps}
\end{figure}

We demonstrate the use of participation maps for visualizing student participation
      in Fig. \ref{fig:ActivityMaps}.
We note that the ground truth plot of Fig. \ref{fig:ActivityMaps}(a)
      suggests that there are long periods where the students are present.
Without DPT, as shown in Fig. \ref{fig:ActivityMaps}(b),
      tracking fails to track the top and bottom students (Javier67p + Kenneth1P).
With DPT, as shown in Fig. \ref{fig:ActivityMaps}(c),
      we see dramatic performance improvements in       
       tracking the top and bottom students (Javier67p + Kenneth1P).
In this example, the overall accuracy improved by 16.4\%.

%\begin{table}[!t]
%\caption{\label{Long_videos_info}Long testing videos information}
%\begin{center}
%\begin{tabular}{ccc}
%    \toprule % booktabs 
    
%   \textbf{Video}  & \textbf{Duration} & \textbf{No. of detected people} \\
%%{p{0.1\textwidth}p{0.1\textwidth}p{0.2\textwidth}} \toprule
%\midrule
%	              {V1} & {23m45s}& {4}  \\
   %                                {V2} &{23m45s}& {2}     \\  
      %                             {V3} & {23m45s}& {4}  \\
         %                          {V4} &{23m45s}& {5}   \\
            %                       {V5} & {23m45s}& {4} \\
               %                    {V6} &{23m45s}& {3}   \\
                                                                                            		
	%		\bottomrule
	%	\end{tabular}
%	\end{center}
%\end{table}

%The input for this group of testing data is from the output of the person recognizer \cite{tran2021facial} developed by our group. 

%Six long videos from different groups are tested. Each video is 23m45s, and the frame rate is 30. The total length of this group of the testing dataset is 2h22m30s. Each group includes 2$\sim$5 people (See Table \ref{Long_videos_info}). The input format of DSC is the same as the testing data type I, including person labels and the information of related bounding boxes. 

%Table \ref{table:occ_rec} shows the results before and after applying DSC to the recognition detector.
%As shown in Table 7, with the DSC system, participation accuracy has significantly improved. From the results, our system gave an average accuracy of 82.3\% compared to 61.9\% for the results without the DSC system.

%%%%%%%%%%%%%%%%%%%%%%%%%%%%%%%%%

\section{Conclusion}\label{sec:conclusion}
The paper describes our efforts to build a system to assess student-participation
       in real-life collaborative learning videos.
The real-life dataset presented many challenges that are not represented in 
      standard occlusion datasets.
We developed a new system to address the unique challenges.
Specifically, we developed methods for student group detection using multiple
     representations, video face recognition, and dynamic participant tracking.
We then document excellent performance by our proposed system 
     that is significantly better than other methods.     
We verify our system on long videos of over 20 minutes and also
     provide visualization using student participation maps.

%%%%%%%%%%%%%%%%%%%%%%%%%%%%%%
\begin{table*}[!t]
\caption{\label{table:occ_det_g1}
              Comparison of DPT versus SORT\_OH \cite{nasseri2021simple} for group 1 (1 out of 5).
              The results are computed over the AOLME-DST dataset.}
%\resizebox{\textwidth}{!}{
\begin{center}
\begin{tabular}{lllllllllllll}
\toprule
\multirow{2}{*}{\phantom{a}}      & \multirow{2}{*}{\textbf{Method}}&\phantom{abc} & \multicolumn{8}{c}{\textbf{Person Label}}                  & \multirow{2}{*}{\textbf{Average}} \\ \cmidrule{4-11}
           &        &                       & \textbf{Kenneth1P}&\phantom{abc}  & \textbf{Jesus69P} &\phantom{abcd} & \textbf{Chaitu} &\phantom{abcde} & \textbf{Javier67P} &\phantom{abcdefg} &                                   \\  \midrule
\multirow{2}{*}{\textbf{V1}}&   \textbf{SORT\_OH} &                  & \textcolor{red}{27.8 \%}        &     & \textcolor{red}{47.6\%}         &   & 93.7\%       &   & 99.7\%         &    & 67.2\%                            \\
                             &  \textbf{Ours}   &                  & 100\%           &   & 81.7\%        &    & 100\%         &  & 100\%           &   & 95.4\%                            \\  \midrule
\multirow{2}{*}{\textbf{V2}}  & \textbf{SORT\_OH}  &                 & 82.7\%         &    & 95.2\%        &    & 100\%      &     & 100\%             & & 94.5\%                            \\
                             & \textbf{Ours}       &              & 100\%        &      & 100\%        &     & 100\%          & & 100\%            &  & 100\%                             \\  \midrule
\multirow{2}{*}{\textbf{V3}}  & \textbf{SORT\_OH}   &                & 75\%      &         & 100\%        &     & 100\%     &      & 100\%       &       & 93.7\%                            \\
                              & \textbf{Ours}              &       & 100\%          &    & 100\%       &      & 100\%        &   & 100\%          &    & 100\%                             \\  \midrule
\multirow{2}{*}{\textbf{V4}}  & \textbf{SORT\_OH}   &                & 83.6\%       &      & 100\%      &       & 100\%      &     & 100\%         &     & 95.9\%                            \\
                             & \textbf{Ours}                   &  & 100\%       &       & 100\%        &     & 100\%     &      & 100\%         &     & 100\%                             \\  \midrule
\multirow{2}{*}{\textbf{V5}}  & \textbf{SORT\_OH}  &                 & 74.5\%           &  & 100\%        &     & 92.1\%    &      & 100\%       &       & 91.6\%                            \\
                              & \textbf{Ours}                     && 100\%         &     & 100\%       &      & 91.4\%       &   & 100\%         &     & 97.8\%                            \\  \midrule
\multirow{2}{*}{\textbf{V6}} & \textbf{SORT\_OH}        &           & 92.4\%      &       & 100\%       &      & 100\%       &    & 100\%        &      & 98.1\%                            \\
                              & \textbf{Ours}                     && 100\%         &     & 100\%         &    & 100\%    &       & 100\%        &      & 100\%                             \\  \midrule
\multirow{2}{*}{\textbf{V7}}  & \textbf{SORT\_OH}   &                & 87.4\%          &   & 100\%        &     & 100\%   &        & 100\%       &       & 96.9\%                            \\
                              & \textbf{Ours}                    & & 100\%       &       & 100\%        &     & 100\%      &     & 100\%        &      & 100\%                             \\  \midrule
\multirow{2}{*}{\textbf{V8}}  & \textbf{SORT\_OH}   &                & 99.2\%        &     & 100\%       &      & 100\%      &     & 100\%         &     & 99.8\%                            \\
                              & \textbf{Ours}                   &  & 100\%          &    & 100\%      &       & 100\%    &       & 100\%        &      & 100\%                             \\  \midrule
\multirow{2}{*}{\textbf{V9}}  & \textbf{SORT\_OH}   &                & 81.8\%     &        & 100\%     &        & 100\%     &      & 100\%       &       & 95.4\%                            \\
                              & \textbf{Ours}                   &  & 100\%       &       & 100\%         &    & 100\%      &     & 100\%     &         & 100\%                             \\  \midrule
\multirow{2}{*}{\textbf{V10}} & \textbf{SORT\_OH}  &                 & \textcolor{red}{50.8\%}       &      & 100\%         &    & 100\%     &      & 100\%     &         & 87.7\%                            \\
                              & \textbf{Ours}                    & & 100\%     &         & 100\%       &      & 100\%    &       & 100\%         &     & 100\%                             \\  \midrule
\multirow{2}{*}{\textbf{V11}} & \textbf{SORT\_OH}    &               & 81.4\%       &      & 100\%      &       & 100\%    &       & 100\%       &       & 95.3\%                            \\
                              & \textbf{Ours}                     && 100\%         &     & 100\%       &      & 100\%     &      & 100\%     &         & 100\%                             \\  \midrule
\multirow{2}{*}{\textbf{V12}} & \textbf{SORT\_OH}   &                & \textcolor{red}{43.6\%}        &     & 100\%      &       & 100\%    &       & 100\%      &        & 85.9\%                            \\
                              & \textbf{Ours}                     && 100\%     &         & 100\%       &      & 100\%     &      & 100\%     &         & 100\%                             \\  \midrule
\multirow{2}{*}{\textbf{V13}} & \textbf{SORT\_OH}   &                & 100\%          &   & 100\%    &        & 86\%       &   & 100\%      &       & 96.5\%                            \\
                              & \textbf{Ours}                     && 100\%      &        & 100\%    &        & 90\%      &     & 100\%        &      & 97.5\%                            \\  \midrule
\multirow{2}{*}{\textbf{V14}} & \textbf{SORT\_OH}    &               & 98.7\%        &     & 80.8\%         &   & 100\%   &        & 100\%       &       & 94.9\%                            \\
                              & \textbf{Ours}                     && 100\%       &       & 100\%        &     & 100\%      &     & 100\%     &         & 100\%                             \\  \midrule
\multirow{2}{*}{\textbf{V15}} & \textbf{SORT\_OH}     &              & 100\%      &        & 90.7\%     &       & 100\%     &      & 100\%     &         & 97.7\%                            \\
                              & \textbf{Ours}                     && 100\%       &       & 100\%       &      & 100\%    &       & 100\%           &   & 100\%                             \\ \bottomrule
\end{tabular}
\end{center}
\end{table*}

\begin{table*}[!b]
\caption{\label{table:occ_det_g2}
              Comparison of DPT versus SORT\_OH \cite{nasseri2021simple} for group 2 (2 out of 5).
              The results are computed over the AOLME-DST dataset.}
%\resizebox{\textwidth}{!}{
\begin{center}
\begin{tabular}{lllclclclclclcl}
\toprule
\multirow{2}{*}{}            & \multirow{2}{*}{\textbf{Method}} & \multicolumn{12}{c}{\textbf{Person Label}}                                                                         & \multirow{2}{*}{\textbf{Average}} \\ \cmidrule{3-14}
                             &                                  & \textbf{Kelly}&\phantom{a} & \textbf{Cindy14P} &\phantom{}& \textbf{Carmen13P}&\phantom{} & \textbf{Marina15P}&\phantom{} & \textbf{Marta12P} &\phantom{}& \textbf{Scott}&\phantom{} &                                   \\  \midrule
\multirow{2}{*}{\textbf{V1}} & \textbf{SORT\_OH}                  & 100\%      &    & 100\%         &    & 100\%           &   & 97.5\%        &     & 100\%        &     & 100\%        &  & 99.6\%                            \\
                             & \textbf{Ours}                    & 100\%      &    & 100\%      &       & 100\%        &      & 100\%            &  & 100\%           &  & 100\%       &   & 100\%                             \\  \midrule
\multirow{2}{*}{\textbf{V2}} & \textbf{SORT\_OH}                  & 100\%       &   & 100\%        &     & 100\%        &      & 81.4\%         &    & 100\%         &    & 100\%     &     & 96.9\%                            \\
                             & \textbf{Ours}                    & 100\%      &    & 100\%         &    & 100\%          &    & 100\%          &    & 100\%       &      & 100\%     &     & 100\%                             \\  \midrule
\multirow{2}{*}{\textbf{V3}} & \textbf{SORT\_OH}                  & 99.2\%      &   & 100\%        &     & 100\%         &     & 100\%         &     & 100\%          &   & 100\%         & & 99.9\%                            \\
                             & \textbf{Ours}                    & 100\%       &   & 100\%          &   & 100\%           &   & 100\%      &        & 100\%   &          & 100\%    &      & 100\%                             \\  \midrule
\multirow{2}{*}{\textbf{V4}} & \textbf{SORT\_OH}                  & 100\%      &    & 100\%        &     & 100\%         &     & \textcolor{red}{62.3\%}         &    & 100\%        &     & 100\%        &  & 93.7\%                            \\
                             & \textbf{Ours}                    & 100\%     &     & 100\%        &     & 100\%      &        & 100\%         &     & 100\%         &    & 100\%     &     & 100\%                             \\  \midrule
\multirow{2}{*}{\textbf{V5}} & \textbf{SORT\_OH}                  & 100\%        &  & 100\%          &   & 100\%        &      & 96.3\%    &         & 100\%         &    & 94.8\%       &  & 98.5\%                            \\
                             & \textbf{Ours}                    & 100\%       &   & 93\%         &     & 100\%        &      & 100\%      &        & 100\%        &     & 99.3\%        & & 98.7\%                            \\  \midrule

\multirow{2}{*}{\textbf{V6}} & \textbf{SORT\_OH}                  & 100\%       &   & 100\%           &  & 100\%          &    & 83.1\%       &      & 100\%       &      & 100\%    &      & 97.2\%                            \\
                             & \textbf{Ours}                    & 100\%       &   & 100\%         &    & 100\%          &    & 100\%          &    & 100\%          &   & 100\%      &    & 100\%                             \\  \midrule
\multirow{2}{*}{\textbf{V7}} & \textbf{SORT\_OH}                  & 100\%       &   & 100\%          &   & 100\%          &    & 85.7\%         &    & 99.9\%        &    & 100\%     &     & 97.6\%                            \\
                             & \textbf{Ours}                    & 100\%        &  & 100\%            & & 100\%         &     & 100\%       &       & 98.5\%        &    & 100\%     &     & 99.7\%                            \\ \bottomrule
%\multirow{2}{*}{\textbf{V8}} & \textbf{SORT\_OH}                  & 0\%            & 0\%               & 0\%                & 94.9\%             & 100\%             & 100\%          & 49.2\%                            \\
%                             & \textbf{Ours}                    & 0\%            & 100\%             & 100\%              & 100\%              & 100\%             & 100\%          & 83.3\%                              \\ \hline
\end{tabular}
\end{center}
\end{table*}

\clearpage

% Please add the following required packages to your document preamble:
% \usepackage{multirow}
\begin{table*}[!t]
\caption{\label{table:occ_det_g3}
              Comparison of DPT versus SORT\_OH \cite{nasseri2021simple} for group 3 (3 out of 5).
              The results are computed over the AOLME-DST dataset.}
%\resizebox{\textwidth}{!}{
\begin{center}
\begin{tabular}{lllclclclclcl}
\toprule
\multirow{2}{*}{}            & \multirow{2}{*}{\textbf{Method}} & \multicolumn{10}{c}{\textbf{Person Label}}                                                          & \multirow{2}{*}{\textbf{Average}} \\ \cmidrule{3-12}
                             &                                  & \textbf{Shelby} &\phantom{abcd} & \textbf{Cindy14P} &\phantom{ab} & \textbf{Cesar61P}  &\phantom{ab}& \textbf{Emily62P}  &\phantom{ab}& \textbf{Mauricio60P} &\phantom{a} &                                   \\  \midrule
\multirow{2}{*}{\textbf{V1}} & \textbf{SORT\_OH}                  & 100\%  &         & 100\%       &      & 94\%  &            & 100\%     &        & 100\%    &            & 98.8\%                            \\
                             & \textbf{Ours}                    & 100\%     &      & 100\%          &   & 100\%           &  & 100\%            & & 100\%             &   & 100\%                             \\  \midrule
\multirow{2}{*}{\textbf{V2}} & \textbf{SORT\_OH}                  & 100\%     &      & 100\%        &     & \textcolor{red}{60.5\%}        &    & 100\%             && 100\%               & & 92.1\%                            \\
                             & \textbf{Ours}                    & 100\%        &   & 100\%        &     & 100\%           &  & 100\%         &    & 100\%         &       & 100\%                             \\  \midrule
\multirow{2}{*}{\textbf{V3}} & \textbf{SORT\_OH}                  & 100\%         &  & 100\%         &    & 76.2\%          &  & 100\%            & & 100\%             &   & 95.2\%                            \\
                             & \textbf{Ours}                    & 100\%          & & 100\%            & & 100\%        &     & 100\%          &   & 100\%            &    & 100\%                             \\  \midrule
\multirow{2}{*}{\textbf{V4}} & \textbf{SORT\_OH}                  & 100\%        &   & 100\%          &   & \textcolor{red}{67.8\%}         &   & 100\%          &   & 100\%           &     & 93.6\%                            \\
                             & \textbf{Ours}                    & 100\%         &  & 100\%         &    & 100\%         &    & 100\%         &    & 100\%           &     & 100\%                             \\  \midrule
\multirow{2}{*}{\textbf{V5}} & \textbf{SORT\_OH}                  & 96.6\%         & & 100\%          &   & 79.2\%       &     & 100\%        &     & 100\%            &    & 95.1\%                            \\
                             & \textbf{Ours}                    & 100\%           && 100\%           &  & 100\%         &    & 100\%        &     & 100\%             &   & 100\%                             \\  \midrule
\multirow{2}{*}{\textbf{V6}} & \textbf{SORT\_OH}                  & 100\%          & & 100\%           &  & \textcolor{red}{45.8\%}           & & 100\%          &   & 100\%             &   & 89.2\%                            \\
                             & \textbf{Ours}                    & 100\%          & & 100\%           &  & 100\%           &  & 100\%            & & 100\%             &   & 100\%                             \\ \bottomrule
\end{tabular}
\end{center}
\end{table*}

%%%%%%%%%%%%%%%%%%%%%%%%%%%%%%%%%%%%%%%%%
% Please add the following required packages to your document preamble:
% \usepackage{multirow}
\begin{table*}[!h]
\caption{\label{table:occ_det_g4}
              Comparison of DPT versus SORT\_OH \cite{nasseri2021simple} for group 4 (4 out of 5).
              The results are computed over the AOLME-DST dataset.}
%\resizebox{\textwidth}{!}{
\begin{center}
\begin{tabular}{lllclclclclcl}
\toprule
\multirow{2}{*}{}            & \multirow{2}{*}{\textbf{Method}} & \multicolumn{10}{c}{\textbf{Person Label}}                                                          & \multirow{2}{*}{\textbf{Average}} \\ \cmidrule{3-12}
                             &                                  & \textbf{Julia7P} &\phantom{abc}& \textbf{Martina64P} &\phantom{ab}& \textbf{Bernard129P} &\phantom{ab}& \textbf{Suzie66P} &\phantom{ab}& \textbf{Issac} &\phantom{abc}&                                   \\  \midrule
\multirow{2}{*}{\textbf{V1}} & \textbf{SORT\_OH}                  & 88\%     &        & \textcolor{red}{28.7\%}          &    & 100\%   &             & \textcolor{red}{54\%}       &       & 93.6\%   &      & 72.9\%                            \\
                             & \textbf{Ours}                    & 86.7\%   &        & 100\%&               & 100\%    &            & \textcolor{red}{58.4\%}      &      & 97.1\%      &   & 88.4\%                            \\ \bottomrule
\end{tabular}
\end{center}
\end{table*}

% Please add the following required packages to your document preamble:
% \usepackage{multirow}
\begin{table*}[!b]
\caption{\label{table:occ_det_g5}
              Comparison of DPT versus SORT\_OH \cite{nasseri2021simple} for group 5 (5 out of 5).
              The results are computed over the AOLME-DST dataset.}
%\resizebox{\textwidth}{!}{%
\begin{center}
\begin{tabular}{lllcllclclclcl}
\toprule
\multirow{2}{*}{}            & \multirow{2}{*}{\textbf{Method}} & \multicolumn{11}{c}{\textbf{Person Label}}                                                                              & \multirow{2}{*}{\textbf{Average}} \\ \cmidrule{3-13}
                             &                                  & \textbf{Irma}&\phantom{} & \textbf{Herminio10P} & \textbf{Juan16P}&\phantom{} & \textbf{Jorge17P}&\phantom{} & \textbf{Emilio25P}&\phantom{} & \textbf{Jacinto51P}&\phantom{} &                                   \\  \midrule
\multirow{2}{*}{\textbf{V1}} & \textbf{SORT\_OH}                  & 95.2\%   &     & 100\%               & 76.4\%       &    & 100\%    &         & 100\%       &       & \textcolor{red}{63.6\%}     &         & 89.2\%                            \\
                             & \textbf{Ours}                    & 92.5\%   &     & 100\%                & 100\%   &         & 100\%      &       & 100\%         &     & 100\%     &          & 98.8\%                            \\  \midrule
\multirow{2}{*}{\textbf{V2}} & \textbf{SORT\_OH}                  & 100\%       & & 100\%&                 100\%  &          & 97.1\%           & & 100\%        &      & 73.4\%      &        & 95.1\%                            \\
                             & \textbf{Ours}                    & 100\%   &      & 100\%                & 100\%       &     & 100\%        &     & 100\%         &     & 100\%       &        & 100\%                             \\  \midrule
\multirow{2}{*}{\textbf{V3}} & \textbf{SORT\_OH}                  & 96.2\%   &     & 100\%                & 100\%            && 79.3\%         &   & 99.9\%       &      & 99.7\%          &    & 95.9\%                            \\
                             & \textbf{Ours}                    & 95.7\%     &   & 100\%                & 99.9\%  &         & 100\%          &   & 100\%       &       & 100\%     &          & 99.3\%                            \\  \midrule
\multirow{2}{*}{\textbf{V4}} & \textbf{SORT\_OH}                  & 93.2\%   &     & 100\%                & 85.2\%    &       & 100\%         &    & 100\%          &    & 77.2\%     &         & 92.6\%                            \\
                             & \textbf{Ours}                    & 91.1\%   &     & 100\%                & 99.9\% &          & 100\%       &      & 100\%    &          & 100\%       &        & 98.5\%                            \\  \midrule
\multirow{2}{*}{\textbf{V5}} & \textbf{SORT\_OH}                  & 100\%    &     & 100\%                & 100\%      &      & 100\%    &         & 100\%       &       & \textcolor{red}{60.1\%}   &           & 93.4\%                            \\
                             & \textbf{Ours}                    & 100\%     &    & 100\%                & 100\%        &    & 100\%        &     & 100\%       &       & 100\%        &       & 100\%                             \\  \midrule
\multirow{2}{*}{\textbf{V6}} & \textbf{SORT\_OH}                  & 97.6\%   &     & 100\%                & \textcolor{red}{68.5\%}      &     & 100\%        &     & 100\%    &          & \textcolor{red}{51.4\%}           &   & 86.3\%                            \\
                             & \textbf{Ours}                    & 93.6\%     &   & 100\%               & 100\%   &         & 100\%      &       & 100\%      &        & 100\%        &       & 98.9\%                            \\ \bottomrule
\end{tabular}
\end{center}
\end{table*}
\clearpage

\begin{comment}
\begin{figure*}[!t]
    \centering
    \subfigure[Tracking with occlusion  (SORT\_OH \cite{nasseri2021simple}) Example 1.]
    {
        \includegraphics[width=0.48\textwidth]{Pictures/C5/other_result_v1_250}
       % \label{fig:first_sub}
    }
    \subfigure[DPT Example 1]
    {
        \includegraphics[width=0.48\textwidth]{Pictures/C5/other_result_v2_434}
        %\label{fig:second_sub}
    }
    \\
    \subfigure[Tracking with occlusion  (SORT\_OH \cite{nasseri2021simple}) Example 2.]
    {
        \includegraphics[width=0.48\textwidth]{Pictures/C5/dsc_result_v1_250}
        %\label{fig:second_sub}
    }
    \subfigure[DPT Example 2]
    {
        \includegraphics[width=0.48\textwidth]{Pictures/C5/dsc_result_v2_434}
       % \label{fig:third_sub}
    }
    \caption{Tracking under occlusion comparison between tracking with occlusion  (SORT\_OH \cite{nasseri2021simple}) and DPT.}
    \label{fig:comp_occ}
\end{figure*}

\end{comment}

%%%%%%%%%%%%%%%%%%%%%%%%%%%%%%%%%%%%%%%%
%%%%%%%%%%%%%%%%%%%%%%%%%%%%%%%%%%%%%%%%

\begin{table}[!t]
\caption{\label{table:occ_rec}
              Final system results over the raw, real-life video dataset
                       of AOLME-DLT.
              The use of DPT provided substantially better results than
              the frame-based results that did not use DPT.              
              The duration of each video is 23 minutes and 45 seconds.}
%\resizebox{\textwidth}{!}{
\begin{center}
\begin{tabular}{llll}
\toprule
\multirow{2}{*}{\phantom{a}}  & \multirow{2}{*}{\textbf{Label}} & \multicolumn{2}{c}{\textbf{Accuracy}} \\ \cmidrule{3-4}
                                              &                                                & \textbf{No DPT} & \textbf{DPT}  \\  \midrule
\multirow{5}{*}{\textbf{V1}}&   \textbf{Chaitanya} &     81.5\%           & 86.8\%            \\
		&   \textbf{Kenneth1P} &    47.3\%           &75.5\%            \\
		&   \textbf{Jesus69P} &      91.3\%           & 90.4\%            \\
		&   \textbf{Javier67P} &      21.2\%           & 53.9\%            \\   \cmidrule{2-4}
		&  Average	             & 60.3\%                 & 76.7\%         \\  \midrule

\multirow{3}{*}{\textbf{V2}}&   \textbf{Phuong} &      53.5\%           & 64.6\%            \\
		&   \textbf{Melly77W} &      97.4\%           & 98.6\%       \\  \cmidrule{2-4}
		&  Average	             & 75.5\%                 & 81.6\%         \\  \midrule

\multirow{5}{*}{\textbf{V3}}&   \textbf{Bernard129P} &      38.0\%           & 77.3\%            \\
		&   \textbf{Julia7P} &      56.7\%           & 81.4\%            \\
		&   \textbf{Martina64P} &      31.6\%           & 60.8\%            \\
		&   \textbf{Suzie66P} &      62.3\%           &89.6\%            \\   \cmidrule{2-4}
		&  Average	             & 47.1\%                 & 77.3\%         \\  \midrule

\multirow{6}{*}{\textbf{V4}}&   \textbf{Herman78W} &      79.7\%           & 87.8\%            \\
		&   \textbf{Laura80W} &      40.4\%           & 61.6\%            \\
		&   \textbf{Lucia81W} &      78.8\%           & 92.2\%            \\
		&   \textbf{Mario130W} &      30.0\%           & 84.6\%            \\
		&   \textbf{Melly77W} &      93.5\%           & 96.6\%            \\   \cmidrule{2-4}
		&  Average	             & 64.5\%                 & 84.6\%         \\  \midrule

\multirow{5}{*}{\textbf{V5}}
		&   \textbf{Herminio10P} &     86.9\%           & 93.2\%            \\
		&   \textbf{Katiana73P} &     86.1\%           & 92.4\%            \\
		&   \textbf{Guillermo72P} &      70.1\%           &88.0\%            \\
		&   \textbf{Beto71P} &     38.9\%           & 84.8\%            \\
		   \cmidrule{2-4}
		&  Average	             & 70.5\%                 & 89.6\%         \\  \midrule

\multirow{4}{*}{\textbf{V6}}&   \textbf{Ivonne} &      80.5\%           & 91.3\%            \\
		&   \textbf{Juanita107P} &      71.0\%           & 90.9\%            \\
		&   \textbf{Katiana73P} &      7.8\%           &70.2\%            \\  \cmidrule{2-4}				&  Average	             & 53.4\%                 & 84.1\%         \\

\bottomrule
\end{tabular}
\end{center}
\end{table}

\bibliographystyle{IEEEtran}
\bibliography{reference}

% Generated by IEEEtran.bst, version: 1.14 (2015/08/26)
\begin{thebibliography}{10}
\providecommand{\url}[1]{#1}
\csname url@samestyle\endcsname
\providecommand{\newblock}{\relax}
\providecommand{\bibinfo}[2]{#2}
\providecommand{\BIBentrySTDinterwordspacing}{\spaceskip=0pt\relax}
\providecommand{\BIBentryALTinterwordstretchfactor}{4}
\providecommand{\BIBentryALTinterwordspacing}{\spaceskip=\fontdimen2\font plus
\BIBentryALTinterwordstretchfactor\fontdimen3\font minus
  \fontdimen4\font\relax}
\providecommand{\BIBforeignlanguage}[2]{{%
\expandafter\ifx\csname l@#1\endcsname\relax
\typeout{** WARNING: IEEEtran.bst: No hyphenation pattern has been}%
\typeout{** loaded for the language `#1'. Using the pattern for}%
\typeout{** the default language instead.}%
\else
\language=\csname l@#1\endcsname
\fi
#2}}
\providecommand{\BIBdecl}{\relax}
\BIBdecl

\bibitem{deng2019arcface}
J.~Deng, J.~Guo, N.~Xue, and S.~Zafeiriou, ``Arcface: Additive angular margin
  loss for deep face recognition,'' in \emph{Proceedings of the IEEE/CVF
  conference on computer vision and pattern recognition}, 2019, pp. 4690--4699.

\bibitem{chen2020augmented}
X.~Chen, X.~Xu, Y.~Yang, H.~Wu, J.~Tang, and J.~Zhao, ``Augmented ship tracking
  under occlusion conditions from maritime surveillance videos,'' \emph{IEEE
  Access}, vol.~8, pp. 42\,884--42\,897, 2020.

\bibitem{dong2016occlusion}
X.~Dong, J.~Shen, D.~Yu, W.~Wang, J.~Liu, and H.~Huang, ``Occlusion-aware
  real-time object tracking,'' \emph{IEEE Transactions on Multimedia}, vol.~19,
  no.~4, pp. 763--771, 2016.

\bibitem{yuan2020scale}
Y.~Yuan, J.~Chu, L.~Leng, J.~Miao, and B.-G. Kim, ``A scale-adaptive
  object-tracking algorithm with occlusion detection,'' \emph{EURASIP Journal
  on Image and Video Processing}, vol. 2020, no.~1, pp. 1--15, 2020.

\bibitem{nasseri2021simple}
M.~H. Nasseri, H.~Moradi, R.~Hosseini, and M.~Babaee, ``Simple online and
  real-time tracking with occlusion handling,'' \emph{arXiv preprint
  arXiv:2103.04147}, 2021.

\bibitem{stadler2021improving}
D.~Stadler and J.~Beyerer, ``Improving multiple pedestrian tracking by track
  management and occlusion handling,'' in \emph{Proceedings of the IEEE/CVF
  conference on computer vision and pattern recognition}, 2021, pp.
  10\,958--10\,967.

\bibitem{shi2016human}
W.~Shi, ``{Human Attention Detection Using AM-FM Representations},'' Master's
  thesis, the University of New Mexico, Albuquerque, New Mexico, 2016.

\bibitem{shi2018robust}
W.~Shi, M.~S. Pattichis, S.~Celed{\'o}n-Pattichis, and C.~L{\'o}pezLeiva,
  ``Robust head detection in collaborative learning environments using am-fm
  representations,'' in \emph{2018 IEEE Southwest Symposium on Image Analysis
  and Interpretation (SSIAI)}.\hskip 1em plus 0.5em minus 0.4em\relax IEEE,
  2018, pp. 1--4.

\bibitem{shi2018dynamic}
------, ``Dynamic group interactions in collaborative learning videos,'' in
  \emph{2018 52nd Asilomar Conference on Signals, Systems, and
  Computers}.\hskip 1em plus 0.5em minus 0.4em\relax IEEE, 2018, pp.
  1528--1531.

\bibitem{shi2021person}
------, ``Person detection in collaborative group learning environments using
  multiple representations,'' in \emph{55th Asilomar Conference on Signals,
  Systems, and Computers}.\hskip 1em plus 0.5em minus 0.4em\relax IEEE, 2021,
  pp. 1109--1112.

\bibitem{tran2021facial}
P.~Tran, M.~Pattichis, S.~Celed{\'o}n-Pattichis, and C.~L{\'o}pezLeiva,
  ``Facial recognition in collaborative learning videos,'' in \emph{Computer
  Analysis of Images and Patterns: 19th International Conference, CAIP 2021,
  Virtual Event, September 28--30, 2021, Proceedings, Part II}.\hskip 1em plus
  0.5em minus 0.4em\relax Springer, 2021, pp. 252--261.

\bibitem{jatla2023long}
V.~Jatla, ``Long-term human video activity quantification in collaborative
  learning environments,'' Ph.D. dissertation, The University of New Mexico,
  2023.

\bibitem{jatla2021long}
V.~Jatla, S.~Teeparthi, M.~S. Pattichis, S.~Celed{\'o}n-Pattichis, and
  C.~L{\'o}pezLeiva, ``Long-term human video activity quantification of student
  participation,'' in \emph{2021 55th Asilomar Conference on Signals, Systems,
  and Computers}.\hskip 1em plus 0.5em minus 0.4em\relax IEEE, 2021, pp.
  1132--1135.

\bibitem{teeparthi2021fast}
S.~Teeparthi, V.~Jatla, M.~S. Pattichis, S.~Celed{\'o}n-Pattichis, and
  C.~L{\'o}pezLeiva, ``Fast hand detection in collaborative learning
  environments,'' in \emph{International Conference on Computer Analysis of
  Images and Patterns}.\hskip 1em plus 0.5em minus 0.4em\relax Springer, 2021,
  pp. 445--454.

\bibitem{MOT16}
\BIBentryALTinterwordspacing
A.~Milan, L.~Leal-Taix\'{e}, I.~Reid, S.~Roth, and K.~Schindler, ``{MOT}16: {A}
  benchmark for multi-object tracking,'' \emph{arXiv:1603.00831 [cs]}, Mar.
  2016, arXiv: 1603.00831. [Online]. Available:
  \url{http://arxiv.org/abs/1603.00831}
\BIBentrySTDinterwordspacing

\bibitem{WuLimYang13}
Y.~Wu, J.~Lim, and M.-H. Yang, ``Online object tracking: A benchmark,'' in
  \emph{IEEE Conference on Computer Vision and Pattern Recognition (CVPR)},
  2013.

\bibitem{VOT_TPAMI}
M.~Kristan, J.~Matas, A.~Leonardis, T.~Vojir, R.~Pflugfelder, G.~Fernandez,
  G.~Nebehay, F.~Porikli, and L.~\v{C}ehovin, ``A novel performance evaluation
  methodology for single-target trackers,'' \emph{IEEE Transactions on Pattern
  Analysis and Machine Intelligence}, vol.~38, no.~11, pp. 2137--2155, Nov
  2016.

\bibitem{fan2019lasot}
H.~Fan, L.~Lin, F.~Yang, P.~Chu, G.~Deng, S.~Yu, H.~Bai, Y.~Xu, C.~Liao, and
  H.~Ling, ``Lasot: A high-quality benchmark for large-scale single object
  tracking,'' in \emph{Proceedings of the IEEE/CVF conference on computer
  vision and pattern recognition}, 2019, pp. 5374--5383.

\bibitem{dave2020tao}
A.~Dave, T.~Khurana, P.~Tokmakov, C.~Schmid, and D.~Ramanan, ``Tao: A
  large-scale benchmark for tracking any object,'' in \emph{European conference
  on computer vision}.\hskip 1em plus 0.5em minus 0.4em\relax Springer, 2020,
  pp. 436--454.

\bibitem{lan2021robust}
S.~Lan, J.~Li, S.~Sun, X.~Lai, and W.~Wang, ``Robust visual object tracking
  with spatiotemporal regularisation and discriminative occlusion
  deformation,'' in \emph{2021 IEEE International Conference on Image
  Processing (ICIP)}.\hskip 1em plus 0.5em minus 0.4em\relax IEEE, 2021, pp.
  1879--1883.

\bibitem{kuipers2020hard}
T.~P. Kuipers, D.~Arya, and D.~K. Gupta, ``Hard occlusions in visual object
  tracking,'' in \emph{Computer Vision--ECCV 2020 Workshops: Glasgow, UK,
  August 23--28, 2020, Proceedings, Part V 16}.\hskip 1em plus 0.5em minus
  0.4em\relax Springer, 2020, pp. 299--314.

\bibitem{lecun1998gradient}
Y.~LeCun, L.~Bottou, Y.~Bengio, and P.~Haffner, ``Gradient-based learning
  applied to document recognition,'' \emph{Proceedings of the IEEE}, vol.~86,
  no.~11, pp. 2278--2324, 1998.

\bibitem{guo2021sample}
J.~Guo, J.~Deng, A.~Lattas, and S.~Zafeiriou, ``Sample and computation
  redistribution for efficient face detection,'' \emph{arXiv preprint
  arXiv:2105.04714}, 2021.

\end{thebibliography}

\begin{IEEEbiography}[{\includegraphics[width=1in,height=1.25in,clip,keepaspectratio]{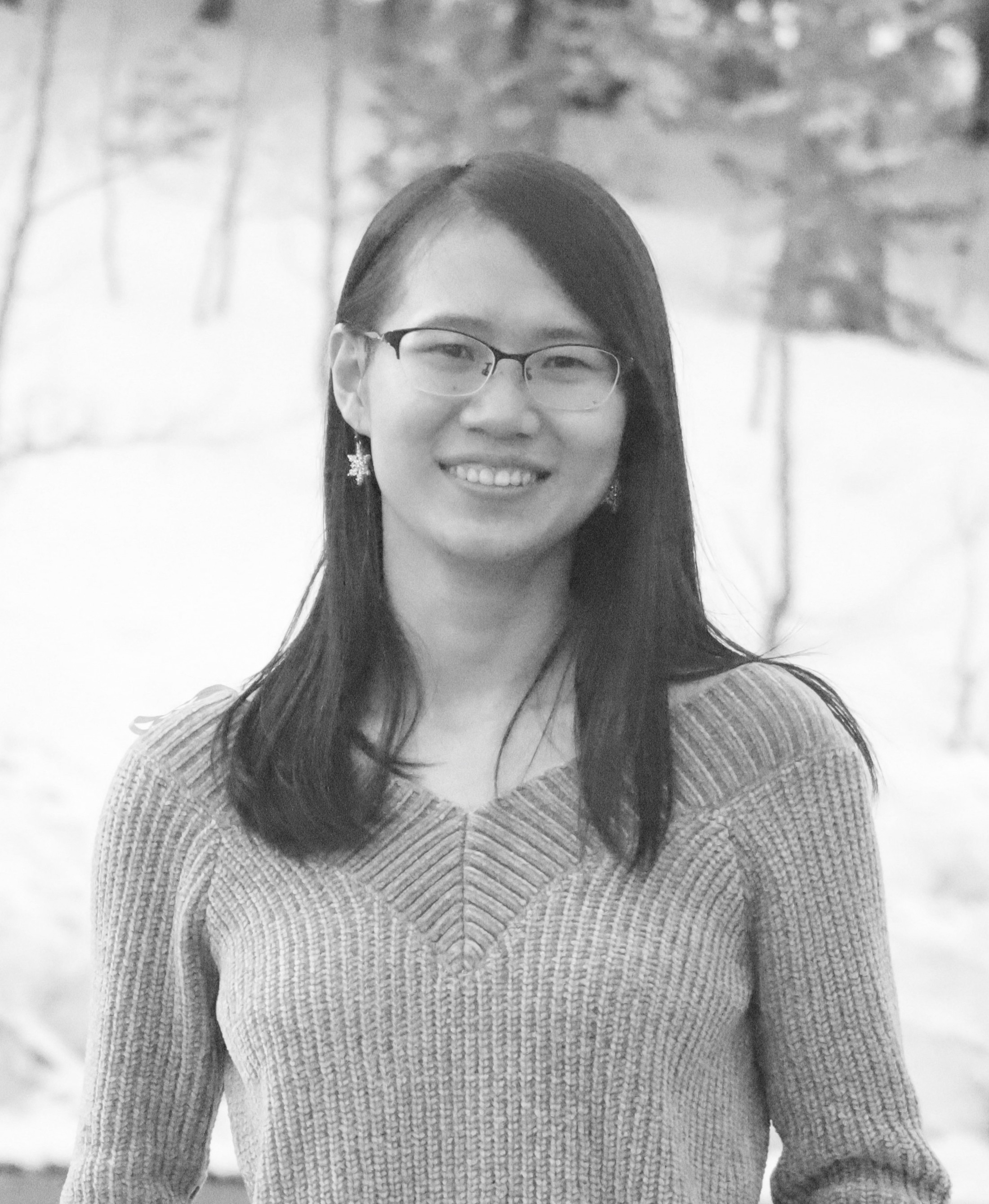}}]{Wenjing Shi} received her M.S. degree in electrical engineering from the University of New Mexico in 2016.
She received her Ph.D. in computer engineering with distinction from the University of the New Mexico in 2023.

Starting from 2017, she has worked as a research assistant at the 
    Image and Video Processing and Communication Lab (ivPCL) and has
    worked as a student facilitator and curriculum developer for the 
    Advancing Out-of-school Learning in Mathematics and Engineering (AOLME) project. Her primary research interests are in video analysis and the development of AM-FM representations and their applications in pose estimation, human detection, and activity detection. Furthermore, in 2021, she worked as a research assistant at the Mind Research Network (MRN), where her focus was on emotion detection. 
Lastly, during the summer of 2022, she pursued an opportunity as an applied scientist intern at Amazon Web Services (AWS) to specialize in image semantic segmentation. She is currently working for Amazon.
\end{IEEEbiography}

\begin{IEEEbiography}
[{\includegraphics[width=1in,height=1.25in, clip, keepaspectratio]{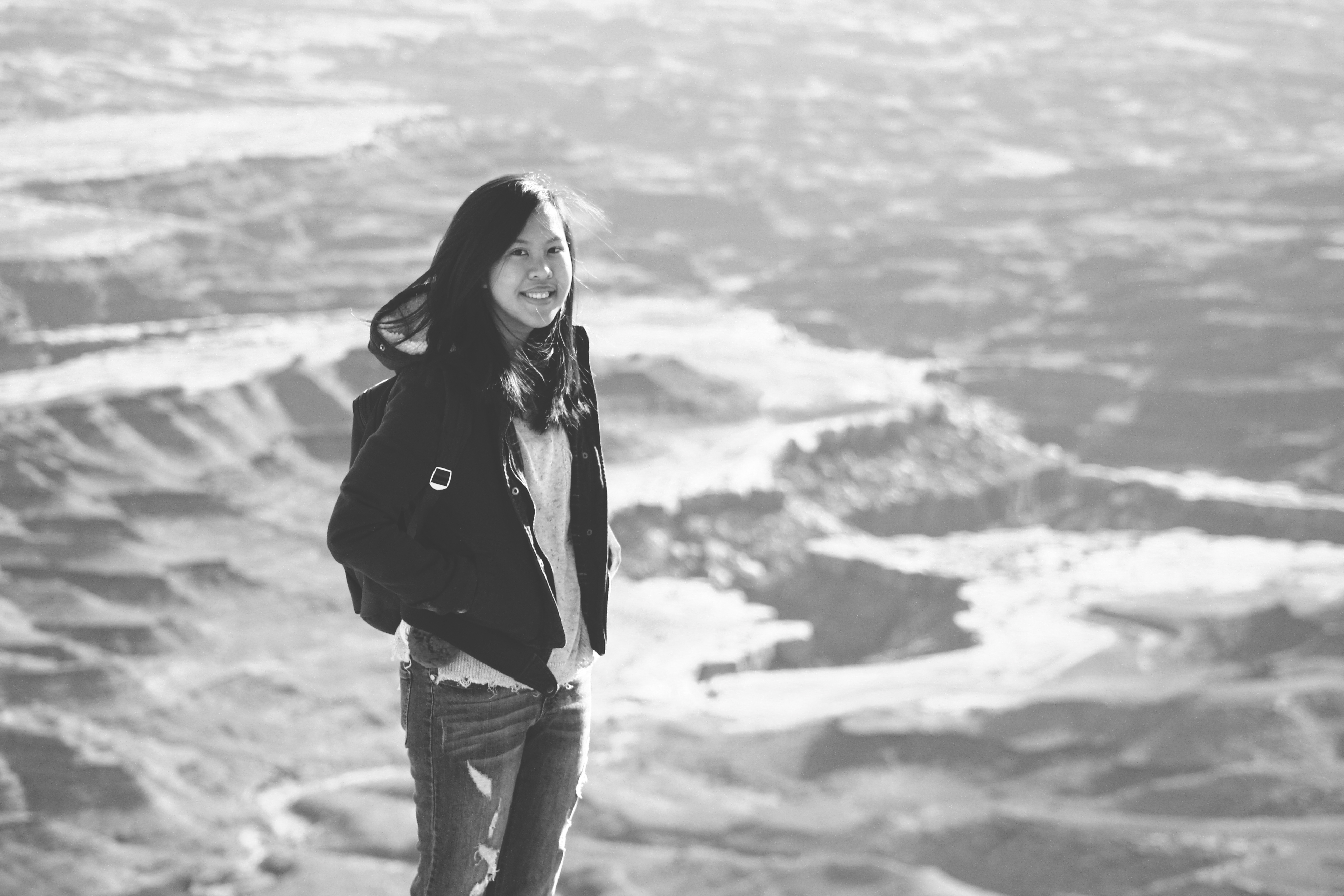}}]{Phuong Tran} received her M.S. degree in computer engineering from the University of New Mexico, Albuquerque, USA, in 2021 and is pursuing a Ph.D. in computer engineering started since August 2022.  

Phuong Tran started as a volunteer teaching middle school students in the AOLME project and later joined as the Research Assistant in the Image and Video Processing and Communication Lab (ivPCL). Her research focuses on detecting and recognizing activities and humans for non-surveillance to assist educational researchers with analyzing students' performance and attendance. In addition to working in multiple National Science Foundations (NSF) projects, she has been an instructor for the Fundamental Programming class at the University of New Mexico. 
\end{IEEEbiography}

\begin{IEEEbiography}
[{\includegraphics[width=1in,height=1.25in,clip,keepaspectratio]{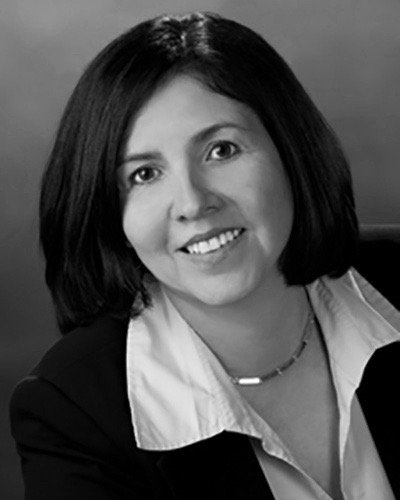}}]{Sylvia Celed\'on-Pattichis} is a professor of bilingual/bicultural education in the Department of Curriculum and Instruction. 
She currently holds the title of H.E. Heartfelder/The Southland Corporation
        Regents Chair in Human Resource Development (Fellow).

Celed\'on-Pattichis prepares elementary pre-service teachers in the bilingual/ESL cohort to teach mathematics and teaches graduate level courses in bilingual education. She taught mathematics at Rio Grande City High School in Rio Grande City, Texas for four years. Her research interests focus on studying linguistic and cultural influences on the teaching and learning of mathematics, particularly with bilingual students. She was a co-principal investigator (PI) of the National Science Foundation (NSF)-funded Center for the Mathematics Education of Latinos/as (CEMELA). She is currently a lead-PI or co-PI of three NSF-funded projects that broaden the participation of Latinx students in mathematics and computer programming in rural and urban contexts.

She serves as a National Advisory Board member of several NSF-funded projects and as an Editorial Board member of the Bilingual Research Journal, Journal of Latinos and Education and Teachers College Record. 
Her current work is a special issue on Teaching and Learning Mathematics and Computing in Multilingual Contexts through Teachers College Record.
She co-edited three books published by the National Council of Teachers of Mathematics titled Access and Equity: Promoting High Quality Mathematics in Grades PreK-2 and Grades 3-5 and Beyond Good Teaching: Advancing Mathematics Education for ELLs.

Celed\'on-Pattichis was a recipient of the Innovation in Research on Diversity in Teacher Education Award from the American Educational Research Association, and the 2011 Senior Scholar Reviewer Award from the National Association of Bilingual Education. She was also a recipient of the Regents Lectureship Award, the Faculty of Color Research Award, Chester C. Travelstead Endowed Faculty Award, and the Faculty of Color Mentoring Award to recognize her research, teaching, and service at The University of New Mexico. 
\end{IEEEbiography}

\begin{IEEEbiography}
[{\includegraphics[width=1in,height=1.25in,clip,keepaspectratio]{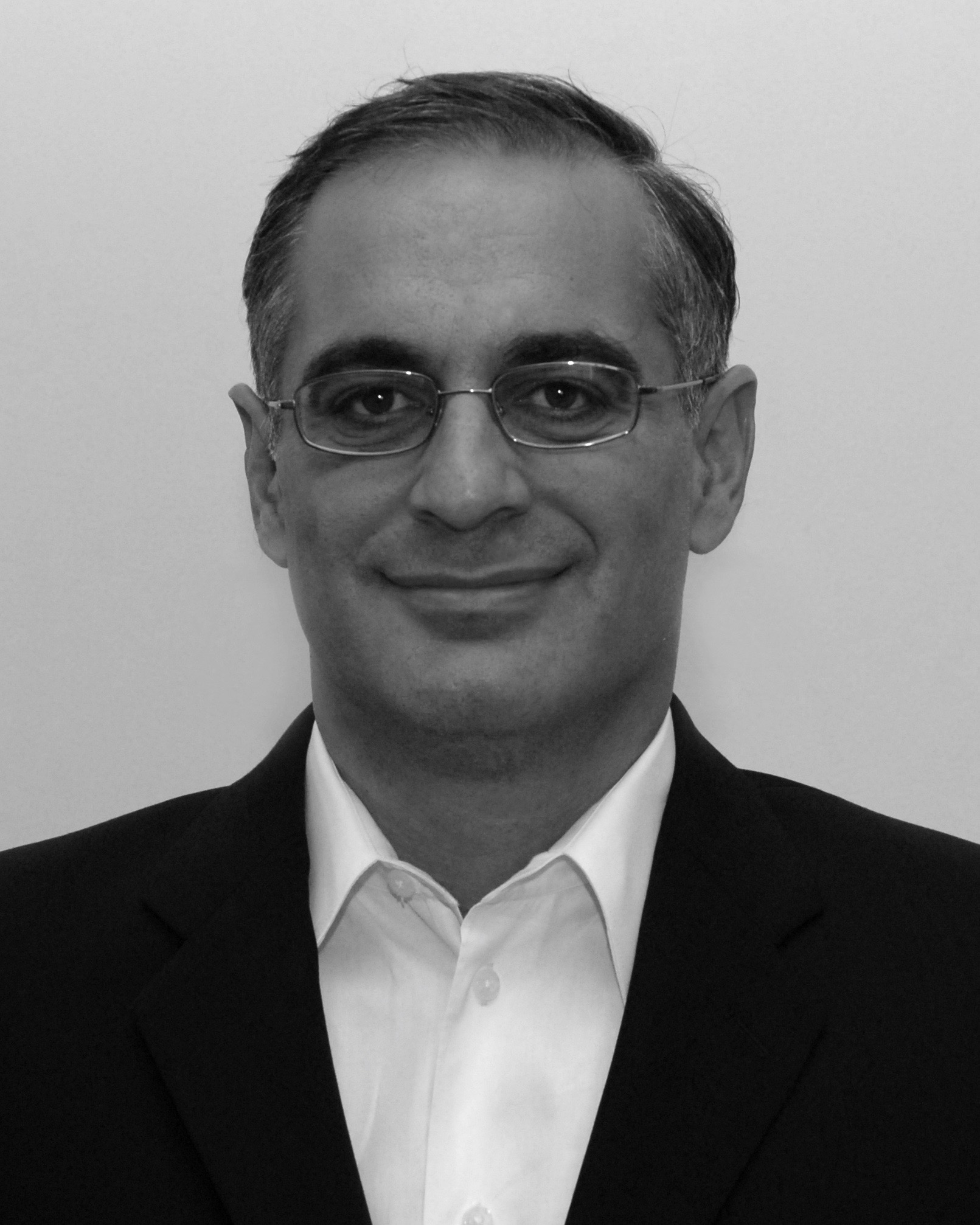}}]{Marios S. Pattichis} received a B.Sc. degree (High Hons. and Special Hons.) in computer sciences, a B.A. degree (High Hons.) in mathematics, an M.S. degree in electrical engineering, and a Ph.D. degree in computer engineering from The University of Texas at Austin, Austin, TX, USA, in 1991, 1991, 1993, and 1998, respectively. 

He is currently a Professor and Director of online programs with the Department of Electrical and Computer Engineering at the University of New Mexico. At UNM, he is also the Director of the Image and Video Processing and Communications Lab (ivPCL). His current research interests include digital image and video processing, video communications, dynamically reconfigurable hardware architectures, biomedical and space image-processing applications, and engineering education.

Dr. Pattichis was a fellow of the Center for Collaborative Research and
   Community Engagement, UNM College of Education, from 2019 to 2020.
He was a recipient of the 2016 Lawton-Ellis and the 2004 Distinguished 
   Teaching Awards from the Department of Electrical and Computer Engineering, UNM. 
For his development of the digital logic design laboratories with UNM, 
    he was recognized by Xilinx Corporation in 2003.
He was also recognized with the UNM School of Engineering’s 
     Harrison Faculty Excellence Award, in 2006. 
     
He was the general chair of the 2008 IEEE Southwest Symposium on 
    Image Analysis and Interpretation (SSIAI), general co-chair of the SSIAI, in 2020 and 2024.
He was also a general chair of the 20th Conference on Computer Analysis of Images and Patterns in 2023.     
He has served as a Senior Associate Editor for the 
    IEEE Transactions on Image Processing 
    and IEEE Signal Processing Letters, an Associate Editor for 
    IEEE Transactions on Image Processing,
    IEEE Transactions on Industrial Informatics, and Pattern Recognition.
He has also served as a Guest Associate Editor for special issues published in the
    IEEE Transactions on Information Technology in Biomedicine,
    Teachers College Record, 
    IEEE Journal of Biomedical and Health Informatics,
    and Biomedical Signal Processing and Control.
He was elected as a fellow of the European Alliance of Medical and
   Biological Engineering and Science (EAMBES) for his contributions to biomedical image analysis.
He was also elected as a Senior Member of the National Academy of Inventors.   
\end{IEEEbiography}
\EOD

\end{document}